\documentclass[10pt,twocolumn,letterpaper]{article}

\usepackage{cvpr}

\AddToShipoutPicture{%
  \AtTextUpperLeft{%
    \put(0,\LenToUnit{1cm}){\parbox{\textwidth}{\centering\color[rgb]{.5,.5,1}\cvprtenhv Preprint}}%
  }%
}

\usepackage{booktabs,graphicx,pifont}
\usepackage[table]{xcolor}
\definecolor{vadHeader}{RGB}{220,230,242}
\definecolor{vadOurs}{RGB}{234,242,250}
\definecolor{vadAP}{RGB}{255,248,218}
\definecolor{vadBest}{RGB}{213,237,211}
\newcommand{\vadYes}{\textcolor{green!55!black}{\ding{51}}}
\newcommand{\vadNo}{\textcolor{red!85!black}{\ding{55}}}
\usepackage{booktabs}
\usepackage{graphicx}
\usepackage[table]{xcolor}
\usepackage{booktabs}
\usepackage{multirow}

\renewcommand{\textfraction}{0.05}

\definecolor{vadHeader}{HTML}{DCE6F1}
\definecolor{vadOurs}{HTML}{EAF1F8}
\definecolor{vadGain}{HTML}{008000}

\usepackage{microtype}
\usepackage{amsmath,amssymb}

\definecolor{cvprblue}{rgb}{0.21,0.49,0.74}
\usepackage[
  pagebackref,
  breaklinks,
  colorlinks,
  allcolors=cvprblue
]{hyperref}

\makeatletter
\newcommand{\nervecaptionof}[1]{\def\@captype{#1}\caption}
\makeatother

\def\confName{CVPR}
\def\confYear{2027}

\title{Weak Labels Suffice to Read Out Event Response Patterns\\
from Frozen MLLM Neurons}

\author{
Pengfei Wei\textsuperscript{1} \quad Sirui Chen\textsuperscript{1} \quad Kei Yiu Kong\textsuperscript{1} \quad Liang Xu\textsuperscript{1} \quad Benfeng Wang\textsuperscript{1} \quad Chao Huang\textsuperscript{1,*}\\
\textsuperscript{1}Sun Yat-sen University\\
{\tt\small \{weipf, chensr63, kongky, xuliang56, wangbf23\}@mail2.sysu.edu.cn, huangch253@mail.sysu.edu.cn}\\
{\small \textsuperscript{*}Corresponding author}
}

\begin{document}

\twocolumn[{%
  \renewcommand\twocolumn[1][]{#1}%
  \maketitle

  \begingroup
    \centering
    \includegraphics[width=\linewidth]{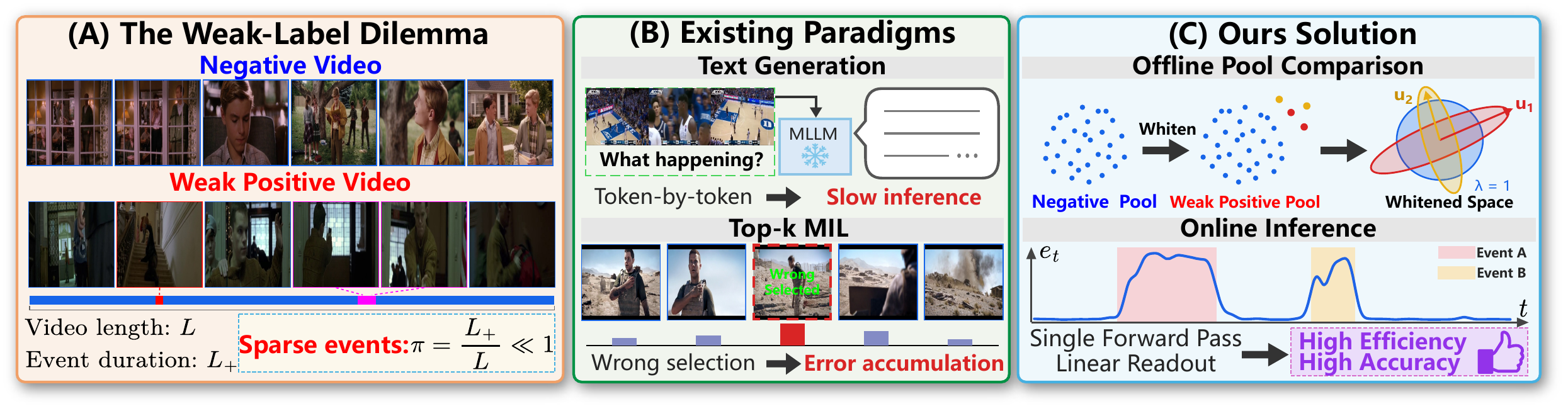}\par

    \vspace{-6pt}

    \nervecaptionof{figure}{%
      \textbf{Weak-label event localization and the idea of NERVE.}
      (a) An event of length $L_+$ occupies only a small fraction
      $\pi=L_+/L$ of a video of length $L$, and supervision is given
      only at the video level.
      (b) Existing approaches compensate for missing supervision
      at the clip level: mis-selected clips reinforce themselves
      in multiple instance learning, and MLLM text generation
      decodes token by token.
      (c) NERVE compares pools rather than clips. After whitening
      the weak positive pool with negative-pool statistics, the
      background becomes a unit circle ($\lambda=1$), events A
      and B are stretched along $\mathbf{u}_1$ and $\mathbf{u}_2$,
      and $\pi$ only changes the amount of stretching.
      One forward pass and one linear projection per clip yield
      the event score.
    }
    \label{fig:teaser}
    \par
  \endgroup

  \vspace{4pt}
}]

\begin{abstract}
Video event localization usually has only video-level labels, while events occupy a small
fraction of each positive video.
Existing methods either iterate on clip pseudo-labels, which accumulates errors, or rely on
autoregressive MLLM generation, which is slow and often requires fine-tuning.
We find that when events occur, the joint activations of frozen MLLM neurons exhibit
low-dimensional event response patterns.
After whitening with reference statistics, background dilution rescales but does not rotate
the event directions, so video-level labels suffice to recover them.
We propose NERVE (NEuron Readout of Video Events), which reads neuron activations in a single
forward pass and obtains an event subspace in closed form by contrasting pool-level
second-order statistics.
A linear projection then yields event scores, with no pseudo-labels, tuning, or text
generation.
With a 2B backbone, NERVE matches or outperforms much larger MLLM-based methods on video
anomaly detection, generalizes to four other event localization tasks and multiple MLLM
families, and runs in real-time streaming mode.
Scoring uses only the current clip, enabling causal inference without accessing future frames
and keeping the per-clip readout cost independent of video length.
\end{abstract}
\section{Introduction}
\label{sec:intro}

Multimodal large language models (MLLMs) understand complex visual scenes well, yet telling
\emph{whether} an event occurs differs from locating \emph{when} it occurs.
Anomaly monitoring, violence and traffic accident detection, highlight extraction, and
action localization need the latter, but supervision usually stops at video-level labels.
Moreover, an event often spans only a fraction $\pi=L_+/L$ of a positive video
(\cref{fig:teaser}a), so most of a positive video looks like a negative one.
The core difficulty is to distinguish clips without knowing which clips are events.

Existing methods address this at the clip level (\cref{fig:teaser}b).
Multiple instance learning selects top-scoring clips or generates pseudo-labels for
self-training~\cite{sultani2018real,tian2021weakly,feng2021mist}, so early mistakes reinforce
themselves.
MLLM-based methods describe or query each clip~\cite{zanella2024harnessing,yang2024follow},
which avoids training but incurs autoregressive latency, and higher accuracy often requires
fine-tuning~\cite{zhang2024holmes}.
We ask whether event signals can instead be read directly from a frozen MLLM, without
guessing clip labels or updating parameters.

We examine MLP neurons in deep decoder layers.
Individual neurons are polysemantic, but their joint activations are structured.
We use \emph{event response patterns} to denote low-dimensional patterns of joint neuron
activation that deviate from a reference (background) distribution when events occur.
Across MLLM families, their excess variance concentrates in a few directions, well above a
label-permutation null.
The key to recovering these directions from weak labels is to compare pools rather than
clips.
The positive pool mixes background with a few events, while the reference pool from
negative videos contains only background.
After reference whitening, the background becomes isotropic and dilution only scales the
excess variance by $\pi$, leaving the event directions unchanged (\cref{fig:teaser}c).

Building on this, we propose NERVE (NEuron Readout of Video Events).
NERVE runs one forward pass per clip, reads neuron activations instead of generating text,
obtains an event subspace offline by generalized eigendecomposition of the two pools, and
scores each test clip online by a single projection.
Its readout aligns with ground-truth intervals on held-out videos and stays close to the
oracle subspace under controlled dilution.
With a 2B backbone and no tuning, NERVE matches or outperforms much larger MLLM-based
methods on three anomaly detection benchmarks and generalizes to other event types and
model families.
Our contributions are as follows.
\begin{itemize}
  \item We reveal \textbf{low-dimensional event response patterns} in frozen MLLM neurons,
  which exist across model families and align with event intervals on held-out videos.
  \item We show that, after reference whitening, background dilution rescales but does not
  rotate the event directions, and build on this \textbf{weak-label recoverability} to
  propose \textbf{NERVE}, a closed-form readout free of pseudo-labels, gradient updates, and
  text generation.
  \item NERVE is effective on anomaly, violence, and traffic anomaly detection, highlight
  detection, and temporal action localization across MLLM families, and supports
  real-time \textbf{streaming inference}.
\end{itemize}
\section{Related Work}
\label{sec:related}

\paragraph{Weakly supervised video event localization and MLLMs.}
Weakly supervised video anomaly detection (VAD) commonly uses multiple instance learning with
ranking losses~\cite{sultani2018real}, later improved by noisy-label
cleaning~\cite{zhong2019graph}, feature magnitudes~\cite{tian2021weakly}, pseudo-label
self-training~\cite{feng2021mist}, and audio cues~\cite{wu2020not}.
Weakly supervised temporal action localization similarly aggregates clip-level classifiers
into video-level predictions~\cite{wang2017untrimmednets,paul2018wtalc}, and CLIP priors have
been introduced into both~\cite{joo2023clip,wu2024vadclip}.
All of them must decide which clips are events.
MLLM-based methods either output anomaly judgments through instruction
tuning~\cite{zhang2024holmes,lv2024video} or reason in a training-free manner via captioning
and LLM scoring~\cite{zanella2024harnessing}, rule induction~\cite{yang2024follow}, or guiding
questions~\cite{ye2025vera,lrpo2025}, so their localization signals come from generated text
or token probabilities.
Closest to ours are methods using internal representations~\cite{prism2025}, which train
probes, select attention heads, or steer hidden
states~\cite{hiprobevad2025,headhuntvad2025,steervad2025}.
NERVE shares the weak supervision setting but casts localization as second-order contrast
between pools: it uses video-level labels only to fit closed-form statistics, never decodes
text, and neither trains probes nor modifies the model.
This avoids iterative clip selection within positive videos, where sparse events are mixed
with background.

\paragraph{Neuron analysis and contrastive subspace readout.}
MLP layers act as key--value memories~\cite{geva2021transformer}, neuron polysemanticity is
explained by superposition~\cite{elhage2022toy}, and multimodal neurons respond selectively
to visual concepts~\cite{goh2021multimodal}.
Linear probes and representation engineering show that high-level attributes lie along a few
linear directions~\cite{alain2017understanding,zou2023representation}, while contrastive
PCA~\cite{abid2018exploring} and generalized eigendecomposition (GED)~\cite{cohen2022tutorial}
find directions along which one dataset varies more than another.
These tools support interpretation, editing, and exploratory analysis, but do not directly
address whether readout directions survive when positives are diluted by irrelevant samples.
In weakly labeled videos, even the positive pool can be dominated by background,
making event-specific variation difficult to separate from shared nuisance variation.
We adopt GED as a classical tool; our contribution is to show that event subspaces in frozen
MLLM neurons remain recoverable under weak-label dilution and can serve as temporal
localization signals without requiring clip-level annotations.
\section{Motivating Findings}
\label{sec:findings}

We ask whether frozen MLLM neurons carry a readable event signal (\cref{sec:finding1}),
and whether video-level labels alone can recover it (\cref{sec:finding2}).
The protocol matches NERVE: the backbone is frozen, and the readout layer, \texttt{[Last]}
token, input template, and clip stride are fixed (Appendix~A).
All statistics use only video-level training labels; test-set temporal annotations are used
only for evaluation and coloring.
Let $\mathbf{x}_t$ be the calibrated neuron vector of clip $c_t$ over the candidate set
$\mathcal{C}$. $S_{\mathrm{ref}}$ and $S_+$ denote the second-order moments of the reference
and weak positive pools about the reference mean, and $\tilde S_{\mathrm{ref}}$ is a
regularized estimate of $S_{\mathrm{ref}}$.

\begin{figure*}[t]
  \centering
  \includegraphics[width=\linewidth]{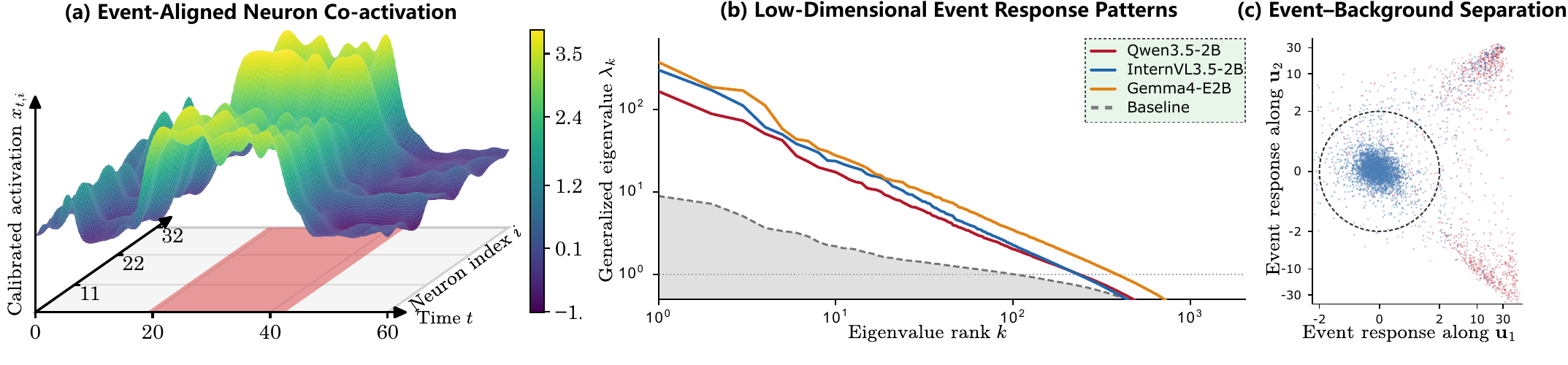}
  \caption{\textbf{Frozen MLLM neurons exhibit low-dimensional event response patterns.}
  (a) Activations of the top-32 neurons, ranked by loading on $\mathbf{u}_1$, on one test
  video (XD-Violence, Qwen3.5-2B, layer 23). Red floor: ground-truth event.
  (b) Generalized eigenspectrum after reference whitening (log--log). Solid: three
  backbones. Dashed with gray fill: label-permutation maximum. Dotted: $\lambda=1$.
  (c) Test clips projected onto the training-set directions $(\mathbf{u}_1,\mathbf{u}_2)$,
  symmetric-log axes. Dashed circle: radius 2. Test labels are used only for coloring.}
  \label{fig:findings}
\end{figure*}

\subsection{Low-Dimensional Event Response Patterns}
\label{sec:finding1}

\paragraph{Event-aligned co-activation.}
Single neurons are unreliable event indicators, but populations are not.
In \cref{fig:findings}(a), neurons ranked by their loadings on the leading direction
$\mathbf{u}_1$ rise together when the event begins and fall together when it ends.
The scene stays largely the same throughout the video, so this response tracks the event
and not a change of scene.
Over the 462 test videos that contain both event and background clips, the mean activation
of the top-32 neurons reaches an average within-video AUC of 80.7\%.
We call such structures event response patterns.

\paragraph{Low-dimensional spectrum.}
We solve $S_+\mathbf{u}=\lambda\,\tilde S_{\mathrm{ref}}\,\mathbf{u}$ and sort the
eigenvalues in descending order:
\begin{equation}
  \lambda_k = \frac{\mathbf{u}_k^\top S_+ \mathbf{u}_k}
                   {\mathbf{u}_k^\top \tilde S_{\mathrm{ref}}\,\mathbf{u}_k},
  \label{eq:rayleigh}
\end{equation}
so $\lambda_k-1$ is the excess variance over background along $\mathbf{u}_k$.
Candidate selection inflates the eigenvalues. To absorb this bias into the null, we permute
video-level labels, rerun the full pipeline including selection, and take the maximum
eigenvalue as the null bound.
In \cref{fig:findings}(b), only the head of the spectrum exceeds this bound on all three
backbones, and the tail quickly decays toward $\lambda=1$.

\begin{table}[t]
\centering
\caption{Generalized eigenspectrum statistics on XD-Violence. Excess variance concentrates
in a few directions, far above the permutation null.}
\label{tab:eigenspectrum}
\begingroup
\setlength{\tabcolsep}{4pt}
\renewcommand{\arraystretch}{1.12}
\setlength{\aboverulesep}{0pt}
\setlength{\belowrulesep}{0pt}
\resizebox{\columnwidth}{!}{%
\begin{tabular}{l|ccc}
\toprule
\rowcolor{vadHeader}
\textbf{Statistic}
& \textbf{Qwen3.5-2B}
& \textbf{InternVL3.5-2B}
& \textbf{Gemma4-E2B} \\
\midrule
Largest eigenvalue $\lambda_1$
& 164.6 & 298.6 & 371.4 \\
Ratio to permutation max
& $22\times$ & $36\times$ & $42\times$ \\
Effective dimension (PR)
& 16.7 & 12.6 & 15.8 \\
\midrule
\multicolumn{4}{l}{%
\cellcolor{white}\textit{Cumulative excess spectral mass (\%)}} \\
\midrule
Top-5   & 46.2 & 51.0 & 46.9 \\
Top-10  & 57.5 & 61.7 & 56.0 \\
Top-32  & 77.4 & 81.1 & 72.3 \\
Top-100 & 94.0 & 95.4 & 88.6 \\
\bottomrule
\end{tabular}%
}
\endgroup
\vspace{2pt}
\parbox{\columnwidth}{\footnotesize
Permutation max is taken over 20 video-level label permutations, each rerun through the
full pipeline. PR follows \cref{eq:pr}; Top-$n$ is $100\sum_{k\le n}w_k/\sum_k w_k$.
}
\end{table}

As \cref{tab:eigenspectrum} shows, $\lambda_1$ is at least $22\times$ the permutation
maximum on every model.
We measure effective dimension by the participation ratio of the excess spectrum:
\begin{equation}
  \mathrm{PR}=\frac{\big(\sum_k w_k\big)^2}{\sum_k w_k^2},\qquad
  w_k=\max(\lambda_k-1,0).
  \label{eq:pr}
\end{equation}
PR is far below $|\mathcal{C}|$, and the top 32 directions hold over 70\% of the excess
mass. This motivates NERVE's default $K$ (\cref{sec:subspace}, \cref{fig:sensK}).
Because $S_+$ comes from a pool dominated by background, this structure already suggests
that weak labels are enough to expose it.

\paragraph{Separability on held-out clips.}
We project test clips onto the training-set directions $(\mathbf{u}_1,\mathbf{u}_2)$.
Since $\mathbf{u}_k^\top\tilde S_{\mathrm{ref}}\mathbf{u}_k=1$, background clips should
have roughly unit variance along each direction.
In \cref{fig:findings}(c), 89.6\% of background clips fall inside the radius-2 circle. This
is close to the $1-e^{-2}\approx86.5\%$ expected for a 2D standard Gaussian, so the whitened
background stays nearly isotropic at test time.
Meanwhile, 82.4\% of event clips fall outside the circle, mostly along $+\mathbf{u}_1$.
This motivates accumulating only positive responses (\cref{sec:scoring}).

In summary, events induce a low-dimensional, structured joint response in frozen MLLM
neurons. It appears in three model families and generalizes to held-out clips.
This does not imply monosemantic neurons, one event category per direction, or neuron
coordinates shared across models.

\subsection{Weak Labels Recover Event Directions}
\label{sec:finding2}

Most clips in the weak positive pool are background. We ask whether this biases the readout
directions, and whether the bias grows as the event fraction shrinks.
Unlike pseudo-labeling, statistical readout never picks clips. It only compares
second-order statistics of two pools.

\paragraph{Idealized population analysis.}
Assume non-event clips in the weak positive pool follow the reference distribution, events
make up a fraction $\pi$, and all moments are taken about the reference mean. Then
\begin{equation}
  S_+ = (1-\pi)\, S_{\mathrm{ref}} + \pi\, S_{\mathrm{event}},
  \label{eq:mixture}
\end{equation}
and whitening gives
\begin{equation}
  S_{\mathrm{ref}}^{-1/2} S_+ S_{\mathrm{ref}}^{-1/2}
  = (1-\pi)\, I + \pi\, S_{\mathrm{ref}}^{-1/2} S_{\mathrm{event}}\, S_{\mathrm{ref}}^{-1/2}.
  \label{eq:whitened_mixture}
\end{equation}
The identity term does not affect the eigenvectors, so they are independent of $\pi$, and
\begin{equation}
  \lambda_k(\pi) = 1 + \pi\,\big(\lambda_k(1) - 1\big),
  \label{eq:linear_law}
\end{equation}
where $\lambda_k(1)$ is the oracle eigenvalue from event clips only.
For any $\pi>0$, eigenvalue order is preserved. The weak-label subspace therefore matches
the oracle subspace, and dilution only shrinks the eigenvalues linearly toward 1.

\paragraph{When the invariance holds.}
The analysis relies on three conditions.
(i) \emph{Distribution consistency}: if positive videos come from systematically different
scenes, scene directions can leak into the readout.
(ii) \emph{A shared coordinate system}: centering each pool on its own mean adds a rank-one
mean-difference term with coefficient $\pi(1-\pi)$. This term rotates the directions as
$\pi$ changes, so NERVE computes all moments about the reference mean.
(iii) \emph{Finite samples}: a smaller $\pi$ shrinks the eigengap and increases estimation
variance.
Regularization and candidate selection make the invariance approximate (Appendix~D).

\begin{figure}[t]
  \centering
  \includegraphics[width=\linewidth]{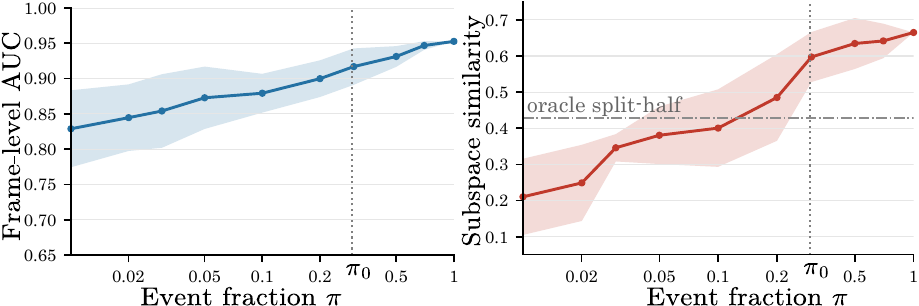}
  \caption{\textbf{Weak-label recovery under controlled dilution} (UBI-Fights).
  x-axis: event fraction $\pi$ (log scale).
  \textbf{Left:} frame-level AUC on split B.
  \textbf{Right:} weak-label vs.\ oracle subspace similarity ($K=4$; dash-dotted: oracle split-half baseline).
  Shaded: $\pm1$ std over resamplings.
  Vertical dotted: natural fraction $\pi_0\approx0.30$.}
  \label{fig:recovery}
\end{figure}

\paragraph{Controlled dilution.}
We split the frame-annotated UBI-Fights training set into disjoint splits A and B.
On split A, we mix event and background clips at a set $\pi$, read out the subspace with
the NERVE pipeline, and evaluate on split B. $\pi=1$ gives the oracle.
In \cref{fig:recovery}, as $\pi$ decreases, split-B AUC (left) and subspace similarity to
the oracle (right; Appendix~D) both decline slowly.
At the natural $\pi_0\approx0.3$, AUC is 91.7\%, versus 95.3\% for the oracle. For $\pi\ge0.3$,
$\lambda_1(\pi)$ stays within 6\% of the linear law.
Over a wide range of event fractions, video-level labels thus recover the event directions
stably. This is why NERVE assigns no clip pseudo-labels.

\section{Method: NERVE}
\label{sec:method}

\begin{figure*}[t]
  \centering
  \includegraphics[width=\linewidth]{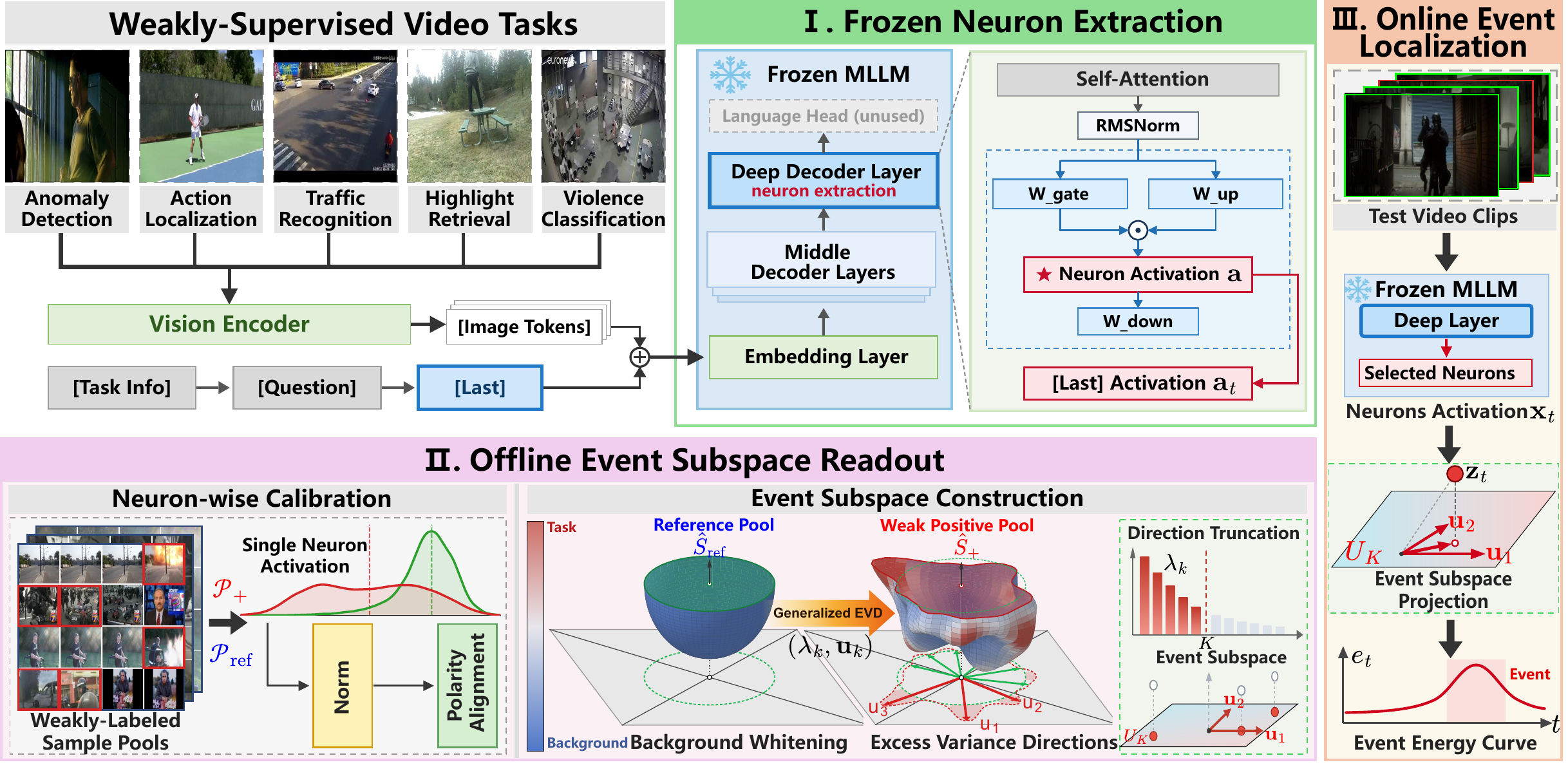}
  \caption{\textbf{Overview of NERVE.}
  (I) Visual tokens of a clip and a shared template pass once through the frozen MLLM;
  neuron activations are read from a deep decoder MLP at the \texttt{[Last]} token.
  (II) Offline, training clips are split by video-level labels into a reference pool and a
  weak positive pool. Candidate neurons are calibrated with reference statistics, and
  reference whitening followed by generalized eigendecomposition yields the event subspace
  $U_K$ and per-direction weights in closed form.
  (III) Online, each test clip is calibrated and projected onto $U_K$, and its one-sided
  weighted energy gives the event energy $e_t$.}
  \label{fig:pipeline}
\end{figure*}

NERVE reads neuron activations from a frozen MLLM in a single forward pass
(\cref{sec:setup}), calibrates them against a reference pool (\cref{sec:calibration}),
obtains the event subspace in closed form offline (\cref{sec:subspace}), and scores new
clips online by a linear projection (\cref{sec:scoring}).

\subsection{Problem Setup and Neuron Readout}
\label{sec:setup}
\label{sec:extraction}

Each training video $v$ has only a video-level label $y_v\in\{0,1\}$, where $y_v=1$
means the target event occurs somewhere in the video. Videos are cut into clips $c_t$ at a
fixed stride. Clips from videos with $y_v=0$ form the reference pool
$\mathcal{P}_{\mathrm{ref}}$; those with $y_v=1$ form the weak positive pool $\mathcal{P}_+$.
The reference pool contains only background, while the weak positive pool mixes background
and event clips. Its event fraction $\pi$ is used only for analysis.

The visual tokens of each clip are concatenated with a shared text template (a task
description, a question, and a final \texttt{[Last]} token) and fed once into the frozen
MLLM. Let $\mathbf{h}_t$ be the normalized hidden state entering the MLP of a selected deep
decoder layer. We read the activations after the gated product and before the
down-projection,
\begin{equation}
  \mathbf{a}_t=\Big[\phi\big(W_{\mathrm{gate}}\mathbf{h}_t\big)\odot
  W_{\mathrm{up}}\mathbf{h}_t\Big]_{[\mathrm{Last}]},
  \label{eq:neuron}
\end{equation}
where $\mathbf{a}_t\in\mathbb{R}^{D}$, $\phi$ is the MLP activation, and the subscript denotes the \texttt{[Last]} position,
the only position that sees all visual tokens and the instruction under causal attention.
Each coordinate of $\mathbf{a}_t$ is a neuron. Under superposition, many sparse features
share this space along nearly orthogonal directions~\cite{elhage2022toy}, so event signals
appear as joint directions over neuron groups (\cref{sec:finding1}). We therefore read out
a subspace rather than individual neurons. Layer and template choices are given in
Appendix~A, the neuron-space geometry in Appendix~E, and a comparison with residual-stream
and attention features in Appendix~B. For temporal action localization, pools are built per
class $j$, and the corresponding quantities carry the superscript $(j)$.

\subsection{Reference Calibration}
\label{sec:calibration}

Raw neuron activations vary widely in scale and sign, and most neurons are unrelated to the
event. Let $\mu_{\mathrm{ref},i},\sigma_{\mathrm{ref},i}$ and $\mu_{+,i},\sigma_{+,i}$ be the
mean and standard deviation of the $i$-th neuron in the two pools. Background dilution
shrinks the between-pool mean gap to $\pi$ times that of the event, but the extra
fluctuation of events still raises the variance of the weak positive pool. We score each
neuron by its variance ratio $\sigma_{+,i}^2/\sigma_{\mathrm{ref},i}^2$ multiplied by
$\sigma_{+,i}$ to account for response magnitude, and keep the top neurons as the candidate
set $\mathcal{C}$ (size in Appendix~A). Each candidate is then centered, scaled, and
polarity-aligned with reference statistics,
\begin{equation}
  x_{t,i}=s_i\,\frac{a_{t,i}-\mu_{\mathrm{ref},i}}{\sigma_{\mathrm{ref},i}},
  \qquad i\in\mathcal{C},
  \label{eq:calib}
\end{equation}
where $s_i=\operatorname{sign}(\mu_{+,i}-\mu_{\mathrm{ref},i})$ aligns each neuron's polarity.
Centering at the reference mean places both pools in a shared background-centered frame,
the premise of the dilution invariance in \cref{sec:finding2}. Scaling by
$\sigma_{\mathrm{ref},i}$ prevents large-scale neurons from dominating the regularization,
and the polarity $s_i$ makes event-induced shifts positive in every coordinate. Both are
invertible diagonal transforms and, ignoring regularization, leave the
eigenvalues unchanged.

\subsection{Closed-Form Event Subspace Readout}
\label{sec:subspace}

Background dilution attenuates the between-pool mean shift, while variation within the
weak positive pool can also reflect background structure. Contrasting second moments
seeks joint directions with excess variation relative to background, without assigning
clip-level pseudo-labels. In the calibrated coordinates, we compute
\begin{equation}
  S_q=\frac{1}{|\mathcal{P}_q|}\sum_{t\in\mathcal{P}_q}
  \mathbf{x}_t\mathbf{x}_t^{\top},\qquad q\in\{\mathrm{ref},+\}.
  \label{eq:moments}
\end{equation}
Neither pool subtracts its own mean, avoiding the $\pi$-dependent rank-one term in
\cref{sec:finding2}. We seek directions of maximal variation relative to the reference:
\begin{equation}
  S_{+}\mathbf{u}_k=\lambda_k\,\tilde S_{\mathrm{ref}}\,\mathbf{u}_k,
  \label{eq:ged}
\end{equation}
where $\tilde S_{\mathrm{ref}}$ is the regularized reference moment (Appendix~A),
$\mathbf{u}_k^{\top}\tilde S_{\mathrm{ref}}\mathbf{u}_k=1$, and
$\lambda_1\ge\lambda_2\ge\cdots$.
We whiten with $W=\tilde S_{\mathrm{ref}}^{-1/2}$, eigendecompose $WS_+W$, and map the
eigenvectors back through $W$. In the ideal whitened coordinates, background has unit
variance, so $\lambda_k-1$ measures excess variation. Under distribution consistency,
the recovered directions match the oracle ones; dilution only shrinks eigenvalues toward~1
(\cref{sec:finding2}; derivation and error analysis in Appendix~D).

The top $K$ directions form $U_K$, with weights $w_k=\max(\lambda_k-1,0)$.
We orient each direction to give the weak positive pool a positive mean projection.
$K$ follows the spectral concentration in \cref{tab:eigenspectrum}
(value in Appendix~A, sensitivity in \cref{fig:sensK}).

\begin{table*}[t]
\centering
\caption{Comparison with VLM- and MLLM-based methods on UCF-Crime,
UBnormal, and XD-Violence. Tuning-free denotes the absence of
task-specific gradient updates to any model component; NERVE uses
video-level training labels only for closed-form statistical fitting.
``--'' denotes unavailable or unverified information.}
\label{tab:vad_comparison}
\setlength{\tabcolsep}{5pt}
\renewcommand{\arraystretch}{1.12}
\setlength{\aboverulesep}{0pt}
\setlength{\belowrulesep}{0pt}
\setlength{\doublerulesep}{0pt}
\resizebox{\textwidth}{!}{%
\begin{tabular}{l|c|c|c|c|c|c|>{\columncolor{vadAP}}c}
\toprule
\rowcolor{vadHeader}
\textbf{Method} &
\textbf{Venue} &
\textbf{Backbone} &
\textbf{Model Size} &
\textbf{Tuning-free} &
\shortstack{\textbf{UCF-Crime}\\\textbf{AUC (\%)}} &
\shortstack{\textbf{UBnormal}\\\textbf{AUC (\%)}} &
\shortstack{\textbf{XD-Violence}\\\textbf{AP (\%)}} \\
\midrule
\multicolumn{8}{l}{\cellcolor{white}\textit{VLM-based and VLM-assisted Methods}} \\
\midrule
CLIP-TSA~\cite{joo2023clip}
& ICIP'23 & CLIP & $\sim$ 0.2B & \vadNo
& 87.58 & -- & 82.19 \\
VadCLIP~\cite{wu2024vadclip}
& AAAI'24 & CLIP & $\sim$ 0.2B & \vadNo
& 88.02 & -- & 84.51 \\
$\pi$-VAD~\cite{majhi2025just}
& CVPR'25 & I3D & $\sim$2.5B & \vadNo
& 90.33 & -- & 85.37 \\
RefineVAD~\cite{lee2026refinevad}
& AAAI'26 & CLIP + InternVideo2.5 & $\sim$8B & \vadNo
& 88.92 & -- & 88.66 \\
PRISM$^{\dagger}$~\cite{prism2025}
& ICML'26 & Qwen3-VL-Embedding & 2B & \vadYes
& 88.85 & -- & 80.31 \\
\midrule
\multicolumn{8}{l}{\cellcolor{white}\textit{MLLM-based Methods}} \\
\midrule
HiProbeVAD~\cite{hiprobevad2025}
& MM'25 & InternVL2.5 & $\sim$8B & \vadNo
& 86.72 & -- & 82.15 \\
HeadHunt-VAD~\cite{headhuntvad2025}
& AAAI'26 & InternVL3 & $\sim$8B & \vadNo
& 87.03 & -- & 82.63 \\
SteerVAD~\cite{steervad2025}
& ICLR'26 & InternVL3 & $\sim$8B & \vadNo
& 87.15 & -- & 83.02 \\
Anom-$\pi$~\cite{anompi2026}
& ICML'26 & Qwen3-VL & 2B & \vadNo
& 84.46 & 78.75 & 72.29 \\
LAVIDA~\cite{dai2026no}
& CVPR'26 & Qwen2-VL + SAM2 & $\sim$9B & \vadNo
& 82.18 & 76.45 & 90.62 \\
LLaVA-1.5~\cite{liu2024improved}
& CVPR'24 & LLaVA-1.5 & $\sim$7B & \vadYes
& 72.84 & 53.71 & 50.26 \\
EventVAD~\cite{eventvad2025}
& MM'25 & ViT + LLM & $\sim$7B & \vadYes
& 82.03 & -- & 64.04 \\
URF$^{\dagger}$~\cite{urf2025}
& NeurIPS'25 & VideoLLaMA3 + Llama3.1 & 7B + 8B & \vadYes
& 84.28 & 69.02 & 68.07 \\
VADTree~\cite{vadtree2025}
& NeurIPS'25 & LLaVA + Qwen & 7B + 14B & \vadYes
& 84.74 & 65.80 & 68.85 \\
PANDA~\cite{panda2025}
& NeurIPS'25 & Qwen2.5-VL + Qwen2.5 & 7B + 72B & \vadYes
& 84.03 & -- & -- \\
VERA~\cite{ye2025vera}
& CVPR'25 & InternVL2 & 8B & \vadYes
& 86.55 & 71.65$^{*}$ & 70.54 \\
LRPO~\cite{lrpo2025}
& ICML'26 & InternVL3.5 & 8B & \vadYes
& 86.59 & -- & 74.09 \\
\midrule
\rowcolor{vadOurs}
\textbf{NERVE (Ours)}
& -- & \textbf{Qwen3.5} & \textbf{2B} & \vadYes
& \textbf{87.53} & \textbf{72.78}
& \cellcolor{vadBest}\textbf{84.24} \\
\bottomrule
\end{tabular}%
}
\vspace{2pt}
\parbox{\textwidth}{\footnotesize
$^{*}$Reported by LRPO with guiding questions transferred from XD-Violence.
$^{\dagger}$Best configuration selected per dataset.
}
\end{table*}

\begin{table*}[t]
\centering
\caption{Results on four additional video benchmarks (\%).
Zero-shot and NERVE use the same frozen Qwen3.5-2B backbone and input.
$\Delta$: absolute gain over zero-shot (points).
Acc21: clip accuracy over 20 actions and background;
Action Top-1: accuracy on ground-truth action clips.
BACC on UBI-Fights and TAD uses the best threshold; TVSum is averaged over 5 folds.}
\label{tab:additional_video}
\begingroup
\setlength{\aboverulesep}{0pt}
\setlength{\belowrulesep}{0pt}
\setlength{\tabcolsep}{4.5pt}
\renewcommand{\arraystretch}{1.25}
\resizebox{\textwidth}{!}{%
\begin{tabular}{l|ccc|ccc|ccc|ccc}
\toprule
\rowcolor{vadHeader}
& \multicolumn{3}{c|}{\textbf{UBI-Fights}}
& \multicolumn{3}{c|}{\textbf{Traffic Anomaly Dataset}}
& \multicolumn{3}{c|}{\textbf{TVSum}}
& \multicolumn{3}{c}{\textbf{THUMOS14}} \\
\rowcolor{vadHeader}
\textbf{Method}
& AUC & AP & BACC
& AUC & AP & BACC
& AUC & AP & BACC
& Frame mAP & Acc21 & Action Top-1 \\
\midrule
Zero-shot
& 88.92 & 32.63 & 82.01
& 71.62 & 41.85 & 74.05
& 66.35 & 64.90 & 62.18
& 14.71 & 7.82 & 18.11 \\
\rowcolor{vadOurs}
\textbf{NERVE (Ours)}
& \textbf{95.76} & \textbf{37.62} & \textbf{92.86}
& \textbf{94.33} & \textbf{58.91} & \textbf{92.76}
& \textbf{76.87} & \textbf{79.80} & \textbf{71.55}
& \textbf{32.00} & \textbf{66.27} & \textbf{75.16} \\
\midrule
\multicolumn{1}{c|}{$\Delta$}
& \textcolor{vadGain}{+6.84}
& \textcolor{vadGain}{+4.99}
& \textcolor{vadGain}{+10.85}
& \textcolor{vadGain}{+22.71}
& \textcolor{vadGain}{+17.06}
& \textcolor{vadGain}{+18.71}
& \textcolor{vadGain}{+10.52}
& \textcolor{vadGain}{+14.90}
& \textcolor{vadGain}{+9.37}
& \textcolor{vadGain}{+17.29}
& \textcolor{vadGain}{+58.45}
& \textcolor{vadGain}{+57.05} \\
\bottomrule
\end{tabular}%
}
\endgroup
\end{table*}

\subsection{Event Energy Scoring and Inference}
\label{sec:scoring}
\label{sec:inference}

A test clip is mapped to $\mathbf{x}_t$ by \cref{eq:neuron,eq:calib}, projected to its event
response $\mathbf{z}_t=U_K^{\top}\mathbf{x}_t$, and scored by the one-sided weighted energy
\begin{equation}
  e_t=\big\|\mathrm{ReLU}(\mathbf{z}_t)\odot\sqrt{\mathbf{w}}\big\|_2.
  \label{eq:energy}
\end{equation}
Here $\mathbf{w}=(w_1,\dots,w_K)$ contains the excess-variance weights.
By the normalization in \cref{eq:ged}, each component of $\mathbf{z}_t$ is measured in units
of background standard deviation (\cref{fig:findings}(c)). We keep only positive responses
because event clips lie mainly on the positive $\mathbf{u}_1$ half-axis, whereas a two-sided
energy would also count background fluctuations. The excess-variance weights give
near-background directions ($\lambda_k\approx1$) small weight, so a larger $K$ does not let
noise directions dominate.

For temporal action localization, the weak positive pool of class $j$ consists of videos
containing that class, and the reference pool of the remaining videos. The same pipeline
yields $U_K^{(j)}$ and per-class energies $e_t^{(j)}$, which are stacked into a single
matrix multiplication. Video-level classes and temporal intervals are obtained by
aggregating energies and grouping consecutive high-energy clips (rules in Appendix~A),
again without any gradient updates.

Scoring depends only on the current clip, so NERVE supports causal streaming. Beyond the
backbone forward pass, it adds only candidate neuron extraction and a low-dimensional
projection; the backbone dominates inference time (\cref{tab:efficiency}, Appendix~A),
with a 0.57\,ms readout.
\section{Experiments}
\label{sec:exp}

\renewcommand{\bottomfraction}{0.6}
\renewcommand{\textfraction}{0.1}
\renewcommand{\topfraction}{0.95}

This section examines the accuracy of NERVE on video anomaly detection
(\cref{sec:exp_vad}), its generalization to broader events
(\cref{sec:exp_broader}), its generality across backbones
(\cref{sec:exp_mllm}), its sensitivity to the subspace dimension $K$
(\cref{sec:exp_k}), its streaming cost (\cref{sec:exp_eff}), and its
event score curves on real videos across the evaluated datasets (\cref{sec:exp_qual}).

\begin{figure*}[t]
  \centering
  \includegraphics[width=\linewidth]{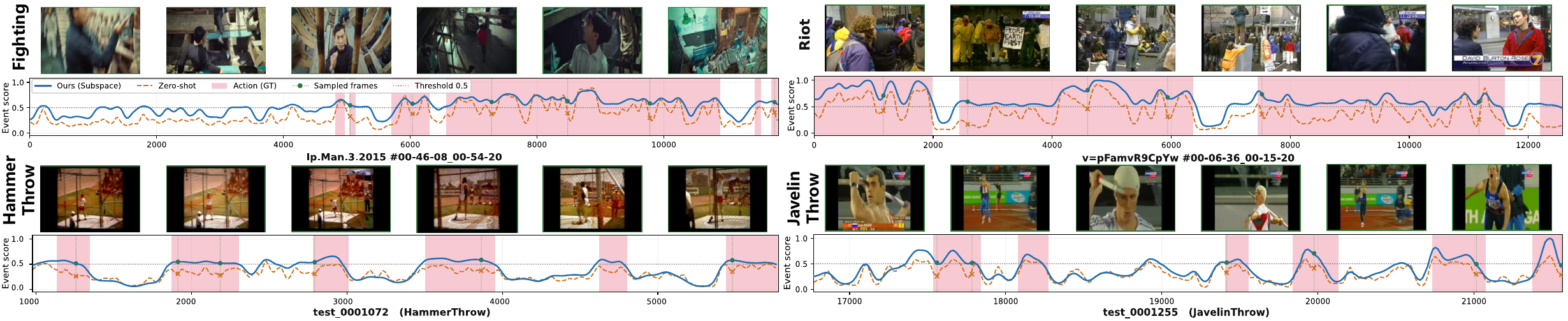}
  \caption{\textbf{Event score $e_t$ of NERVE on test videos from different datasets.}
  Shaded regions mark ground-truth event intervals, used only for visualization.
  Gray lines show the zero-shot $P(\text{yes})$.
  The last panel is a failure case, where abrupt camera motion or a scene change
  that is rare in the reference pool causes a false peak.}
  \label{fig:qual}
\end{figure*}

\subsection{Setup}
\label{sec:exp_setup}

We evaluate NERVE on seven benchmarks: UCF-Crime~\cite{sultani2018real},
UBnormal~\cite{acsintoae2022ubnormal}, and XD-Violence~\cite{wu2020not}
for video anomaly detection; UBI-Fights~\cite{degardin2020iterative}
and TAD~\cite{lv2021localizing} for specific safety events; and
TVSum~\cite{song2015tvsum} and THUMOS14~\cite{jiang2014thumos}
for highlights and multi-class actions.
NERVE uses only video-level training labels.
Dataset splits, label construction, and evaluation protocols are
provided in Appendix~A.

Unless otherwise stated, we use frozen Qwen3.5-2B with a shared
readout configuration and $K=32$ across datasets.
We compare against VLM- and MLLM-based methods in
\cref{tab:vad_comparison}, where tuning-free means no task-specific
gradient updates.
For controlled comparisons, the zero-shot baseline uses the same
frozen backbone and input, with $P(\text{yes})$ as its score.
Additional backbones and efficiency settings are specified in the
corresponding subsections; implementation details are in Appendix~A.

\subsection{Main Results on Video Anomaly Detection}
\label{sec:exp_vad}

We report frame-level AUC on UCF-Crime and UBnormal, and frame-level
AP on XD-Violence in \cref{tab:vad_comparison}.
Unlike most competitors that rely on 7B+ models, multi-model cascades, or
task-specific training, NERVE uses a single frozen 2B backbone and one
shared configuration across all three benchmarks.
On XD-Violence, NERVE reaches 84.24\% AP, the highest among tuning-free
methods, surpassing PRISM by 3.93 points and the strongest tuning-free
generative method LRPO by 10.15 points. Only trained methods score higher:
LAVIDA (90.62\%), RefineVAD (88.66\%), $\pi$-VAD (85.37\%), and VadCLIP
(84.51\%).
On UCF-Crime, it achieves 87.53\% AUC, leading all MLLM-based methods
including those that train probes or steer hidden states (such as
SteerVAD). Among tuning-free methods, only PRISM (88.85\%), which selects
configurations per dataset, is higher. The remaining higher results are from
trained methods: $\pi$-VAD (90.33\%), RefineVAD (88.92\%), and VadCLIP
(88.02\%). CLIP-TSA (87.58\%) is 0.05 points ahead.
On UBnormal, NERVE obtains 72.78\% AUC, again the best among tuning-free
methods and behind only the trained Anom-$\pi$ (78.75\%) and LAVIDA
(76.45\%).
The particularly large gain in XD-Violence AP reflects higher precision
among top-scoring clips.

\subsection{Generalization to Broader Video Events}
\label{sec:exp_broader}

We further evaluate fight detection on UBI-Fights, traffic anomaly
detection on TAD, highlight detection on TVSum, and the multi-class
extension (\cref{sec:scoring}) on THUMOS14.
We report AUC, AP, and balanced accuracy (BACC) on the first three
benchmarks, and frame-level mAP, Acc21, and Action Top-1 on THUMOS14
(\cref{tab:additional_video}; detailed evaluation protocols are provided in Appendix~A).

As shown in \cref{tab:additional_video}, NERVE consistently outperforms
the same-backbone zero-shot baseline across all metrics.
Among binary benchmarks, the AUC gain is largest on TAD (+22.71 points)
and smallest on UBI-Fights, where zero-shot is already strong (88.92\%
AUC), while the improvements on TVSum show that NERVE captures general
deviations from background rather than anomalies alone.
On THUMOS14, because scalar $P(\text{yes})$ is not a multi-class
classifier and stays near chance on 21-way accuracy, the jump in frame
mAP from 14.71\% to 32.00\% serves as the primary indicator of
multi-class localization ability.

\subsection{Generality Across Frozen MLLMs}
\label{sec:exp_mllm}

\begin{table}[!b]
\centering
\caption{Generalization of NERVE across frozen MLLM families (\%).
$\Delta$ denotes absolute improvement over zero-shot in percentage points.}
\label{tab:mllm_generalization}
\begingroup
\setlength{\tabcolsep}{4pt}
\renewcommand{\arraystretch}{1.25}
\setlength{\aboverulesep}{0pt}
\setlength{\belowrulesep}{0pt}
\resizebox{\columnwidth}{!}{%
\begin{tabular}{c|l|cc|c|c}
\toprule
\rowcolor{vadHeader}
& & \multicolumn{2}{c|}{\textbf{XD-Violence}}
& \textbf{UCF-Crime} & \textbf{UBnormal} \\
\rowcolor{vadHeader}
\textbf{Model} & \textbf{Method}
& \textbf{AUC} & \textbf{AP} & \textbf{AUC} & \textbf{AUC} \\
\midrule
& Zero-shot & 91.19 & 73.93 & 83.20 & 69.01 \\
& \cellcolor{vadOurs}\textbf{NERVE}
& \cellcolor{vadOurs}\textbf{95.26}
& \cellcolor{vadOurs}\textbf{84.24}
& \cellcolor{vadOurs}\textbf{87.53}
& \cellcolor{vadOurs}\textbf{72.78} \\
\multirow{-3}{*}{\textbf{Qwen3.5-2B}}
& \multicolumn{1}{c|}{$\Delta$}
& \textcolor{vadGain}{+4.07} & \textcolor{vadGain}{+10.31}
& \textcolor{vadGain}{+4.33} & \textcolor{vadGain}{+3.77} \\
\midrule
& Zero-shot & 89.75 & 73.90 & 80.67 & 70.21 \\
& \cellcolor{vadOurs}\textbf{NERVE}
& \cellcolor{vadOurs}\textbf{95.25}
& \cellcolor{vadOurs}\textbf{84.90}
& \cellcolor{vadOurs}\textbf{86.16}
& \cellcolor{vadOurs}\textbf{72.96} \\
\multirow{-3}{*}{\textbf{InternVL3.5-2B}}
& \multicolumn{1}{c|}{$\Delta$}
& \textcolor{vadGain}{+5.50} & \textcolor{vadGain}{+11.00}
& \textcolor{vadGain}{+5.49} & \textcolor{vadGain}{+2.75} \\
\midrule
& Zero-shot & 85.09 & 62.57 & 74.05 & 64.53 \\
& \cellcolor{vadOurs}\textbf{NERVE}
& \cellcolor{vadOurs}\textbf{90.23}
& \cellcolor{vadOurs}\textbf{70.39}
& \cellcolor{vadOurs}\textbf{79.92}
& \cellcolor{vadOurs}\textbf{68.67} \\
\multirow{-3}{*}{\textbf{Gemma4-E2B}}
& \multicolumn{1}{c|}{$\Delta$}
& \textcolor{vadGain}{+5.14} & \textcolor{vadGain}{+7.82}
& \textcolor{vadGain}{+5.87} & \textcolor{vadGain}{+4.14} \\
\bottomrule
\end{tabular}%
}
\endgroup
\end{table}

To test generality across model families at a similar scale, we
evaluate frozen Qwen3.5-2B, InternVL3.5-2B, and Gemma4-E2B under the
same readout algorithm and configuration ($K=32$), using the readout
layers from \cref{fig:findings}(b) and \cref{tab:eigenspectrum}
(listed in Appendix~A).
As shown in \cref{tab:mllm_generalization}, NERVE improves over
zero-shot across all metrics and backbones, with XD-Violence AP gains
exceeding 10 points on Qwen3.5-2B (+10.31) and InternVL3.5-2B (+11.00),
consistent with their prominent spectral heads above the permutation
null (\cref{tab:eigenspectrum}).
Gemma4-E2B still gains consistently (+5.14 AUC and +7.82 AP on
XD-Violence, +5.87 on UCF-Crime, and +4.14 on UBnormal), although the
improvement is smaller than on the other two backbones, in line with
its lower top-32 spectral mass (72.3\% in \cref{tab:eigenspectrum}).
Furthermore, controlled comparisons in Appendix~B confirm that MLP
neuron activations consistently outperform residual-stream hidden states
and attention outputs under identical pools and readout.

\subsection{Sensitivity to Subspace Dimension $K$}
\label{sec:exp_k}

\begin{figure}[!t]
  \centering
  \includegraphics[width=\linewidth]{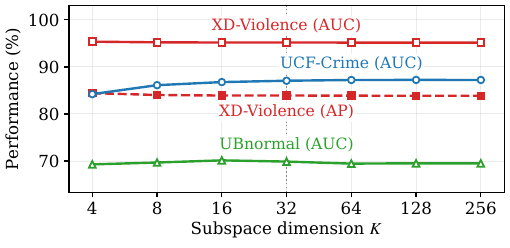}
  \caption{\textbf{Sensitivity to the subspace dimension $K$} (log scale).
  Metrics are AUC and AP on XD-Violence, and AUC on UCF-Crime and UBnormal.
  The dotted line marks the default $K$, chosen from the training-set spectrum.}
  \label{fig:sensK}
\end{figure}

As shown in \cref{fig:sensK}, performance is under-fitted at very small
$K$ but quickly enters a broad plateau across all datasets once $K$
reaches the range implied by the training-set spectral concentration
(\cref{tab:eigenspectrum}).
Increasing $K$ further adds directions with eigenvalues near 1 that
receive negligible weights in \cref{eq:energy}, avoiding performance
degradation.
Accordingly, we fix the default $K=32$ from the training-set spectrum
without test-set tuning; ablations of the remaining components are
provided in Appendix~C.

\subsection{Efficiency and Streaming}
\label{sec:exp_eff}

Measured on a single A100 80GB GPU (batch size 1) against
autoregressive (AR) captioning with the same frozen backbone and input
(\cref{tab:efficiency}), NERVE in streaming mode runs in 245.6\,ms per
clip---$8.4\times$ faster than the 64-token AR lower bound (2062.7\,ms,
excluding downstream LLM scoring)---achieving a $4.07\times$ real-time
factor, reducing time to first score from 3.46\,s to 1.27\,s, and
keeping peak memory unchanged (4.28 vs.\ 4.30\,GB).
The backbone forward pass dominates per-clip latency (94.3\%), whereas
the linear readout takes only 0.57\,ms (0.2\%) and requires 0.28\,MB of
parameters; offline closed-form fitting over 682,773 clips finishes in
16.9\,s using 2.87\,GB of memory without gradient updates or iterative optimization.

\begin{table}[!t]
\centering
\caption{Efficiency and streaming of NERVE on XD-Violence
(frozen Qwen3.5-2B, single A100 80GB).
Clip: 1\,s (24 frames, 4 sampled); RTF = clip duration / latency;
TTFS includes 1\,s buffering.
$^{\dagger}$Captioning only (64 tokens), excluding LLM scoring.}
\label{tab:efficiency}
\begingroup
\small
\setlength{\tabcolsep}{5pt}
\renewcommand{\arraystretch}{1.15}
\setlength{\aboverulesep}{0pt}
\setlength{\belowrulesep}{0pt}
\resizebox{\columnwidth}{!}{%
\begin{tabular}{l|ccc|c}
\toprule
\rowcolor{vadHeader}
\textbf{Inference Mode}
& \shortstack{\textbf{Latency}\\\textbf{(ms)}$\downarrow$}
& \shortstack{\textbf{RTF}\\$(\times)\uparrow$}
& \shortstack{\textbf{TTFS}\\\textbf{(s)}$\downarrow$}
& \shortstack{\textbf{Peak Mem.}\\\textbf{(GB)}$\downarrow$} \\
\midrule
\multicolumn{5}{l}{\cellcolor{white}\textit{(a) End-to-end inference (per 1\,s clip)}} \\
\midrule
AR caption$^{\dagger}$
& 2062.7 & 0.49 & 3.46 & 4.30 \\
\rowcolor{vadOurs}
\textbf{NERVE}
& \textbf{245.6} & \textbf{4.07} & \textbf{1.27} & \textbf{4.28} \\
\midrule
\multicolumn{5}{l}{\cellcolor{white}\textit{(b) Cost breakdown of NERVE}} \\
\midrule
Preprocessing      & \multicolumn{4}{c}{13.3 ms (5.4\%)} \\
Backbone forward   & \multicolumn{4}{c}{231.7 ms (94.3\%)} \\
Readout            & \multicolumn{4}{c}{0.57 ms (0.2\%)} \\
Offline fitting    & \multicolumn{4}{c}{16.9 s} \\
\bottomrule
\end{tabular}%
}
\endgroup
\end{table}

\subsection{Qualitative Analysis}
\label{sec:exp_qual}

\cref{fig:qual} visualizes the event score $e_t$ on test videos across
datasets and event types, where $e_t$ remains flat over background,
rises sharply at event onset, and drops at offset, producing
concentrated peaks even for short events in long videos (the low-$\pi$
regime of \cref{sec:finding2}).
The rightmost panel shows a failure case in which abrupt camera motion
or a scene cut triggers a false peak: because NERVE measures excess
variance relative to the reference distribution rather than high-level
anomaly semantics alone, background states that are rare in the
reference pool deviate from the whitened origin and can project onto
event directions, locally violating the distribution-consistency
condition of \cref{sec:finding2} (more examples in Appendix~E).
\section{Conclusion}
\label{sec:conclusion}

We studied how to read the temporal location of events directly out of a frozen MLLM when
only video-level labels are available.
We found that, when events occur, the joint activations of frozen MLLM neurons exhibit
low-dimensional event response patterns, which exist across multiple model families and
generalize to test clips not used for readout.
Under distribution consistency and a unified coordinate system, background dilution only
shrinks the generalized eigenvalues linearly toward 1 without changing the event
directions, so weak labels recover the event subspace without temporal boundaries or
pseudo-labels.
Controlled dilution experiments show that this property holds approximately under finite
samples and the practical pipeline.
Building on this, NERVE reads out the event subspace in closed form via pool-level
second-order statistical contrast and uses a one-sided weighted energy as the localization
score.
It achieves competitive accuracy on anomaly detection and broader event localization,
generalizes across frozen MLLM backbones, and adds negligible online overhead even for causal streaming inference.

\section*{LLM Usage Statement}
We used Claude (Anthropic) to assist with English language editing and grammar correction,
portions of auxiliary experimental code, and the design of validation experiments.
All LLM-assisted text, code, and experimental designs were reviewed and verified by the
authors, who take full responsibility for the methodology, results, and final content.

{
  \small
  \bibliographystyle{ieeenat_fullname}
  \bibliography{main}
}

\clearpage
\appendix

\section*{Appendix Overview}

This appendix supports the main paper with implementation details, additional
comparisons, theoretical derivations, and further visualizations.

\begin{itemize}\itemsep0pt
\item[\textbf{A.}] \textbf{Implementation details} (\cref{app:impl}): backbones and
readout layer, input template and event slots, candidate selection, regularization and
algorithms, datasets and evaluation protocol, the multi-class extension, comparison
protocol, efficiency measurement, and reproducibility.
\item[\textbf{B.}] \textbf{Readout source} (\cref{app:source}): MLP neuron activations
versus residual-stream hidden states and attention outputs on three backbones.
\item[\textbf{C.}] \textbf{Ablations and sensitivity} (\cref{app:ablation}): component
ablation, alternative weak-label readouts, and sensitivity to $|\mathcal{C}|$ and
$\varepsilon$.
\item[\textbf{D.}] \textbf{Theoretical details} (\cref{app:theory}): the dilution
invariance, per-pool centering, calibration invariance, regularization, finite-sample
behavior, and the case where distribution consistency fails.
\item[\textbf{E.}] \textbf{Visualizations and neuron-space geometry}
(\cref{app:vis}): more activation surfaces, more event score curves, and statistics of
the loading distribution and cross-dataset comparison.
\end{itemize}

\newpage
\section{Implementation Details}
\label{app:impl}

\subsection{Backbones and Readout Layer}
\label{app:backbone}

All three backbones are frozen. NERVE reads the gated-product activation of the last
decoder-layer MLP at the \texttt{[Last]} position (\cref{eq:neuron}). The readout layer is
fixed to the last layer, so no layer is selected on test data (\cref{tab:app_backbone}).
For Qwen3.5-2B this is layer 23, the layer shown in \cref{fig:findings}(a).

\begin{table}[!ht]
\centering
\caption{Backbone and readout configuration. Layers are 0-indexed; $D$ is the MLP
intermediate width.}
\label{tab:app_backbone}
\setlength{\tabcolsep}{4pt}
\renewcommand{\arraystretch}{1.12}
\setlength{\aboverulesep}{0pt}
\setlength{\belowrulesep}{0pt}
\resizebox{\columnwidth}{!}{%
\begin{tabular}{l|cccc}
\toprule
\rowcolor{vadHeader}
\textbf{Backbone} & \textbf{Layers} & \textbf{Readout layer} & $D$ & $K$ \\
\midrule
Qwen3.5-2B     & 24 & last (23) & 6144  & 32 \\
InternVL3.5-2B & 28 & last (27) & 6144  & 32 \\
Gemma4-E2B     & 35 & last (34) & 12288 & 32 \\
\bottomrule
\end{tabular}%
}
\end{table}

$K=32$ is set from the spectral concentration of the training set (the top 32 directions
hold 72.3--81.1\% of the excess spectral mass in \cref{tab:eigenspectrum}). It is shared by
all datasets and backbones and is not tuned on test data (\cref{sec:exp_k}).

\subsection{Input and Template}
\label{app:template}

\paragraph{Clip segmentation.}
Videos are cut into non-overlapping clips with a stride equal to the clip length
(\cref{sec:setup}). The default clip is 1\,s (24 frames), from which 4 frames are sampled
uniformly. On TVSum, a clip is an official 2\,s shot, again with 4 uniformly sampled frames.
On THUMOS14, a clip covers 16 frames, with 4 sampled frames. Input resolution follows each
backbone's default image preprocessing, without extra resizing or cropping. Training and
testing use the same segmentation.

\paragraph{Template.}
The visual tokens of each clip are followed by a shared text template whose last token is
\texttt{[Last]}. Tasks differ only in the event slot \texttt{\{EVENT\}}. The model runs one
forward pass without generating an answer, and NERVE reads the last-layer MLP activation at
\texttt{[Last]} (feature shape $[1, D]$). The prompts were written in Chinese and are shown
here in English translation.

\begin{quote}\small
\textbf{Visual tokens:} four frames of the same clip, uniformly sampled in temporal order.\\
\textbf{[Task]} You are a video analysis assistant. The above are four frames sampled in
temporal order from the same video clip. Using only the visual evidence in the frames,
decide whether the following occurs in the clip: \texttt{\{EVENT\}}.\\
\textbf{[Question]} Based on the temporal changes across the four frames, does the event
above occur in the clip? Answer only ``yes'' or ``no''.\\
\textbf{[Last]} The final token of the input sequence (the start of the assistant turn in
the chat template).
\end{quote}

No decoding is performed after \texttt{[Last]}; all readouts of all tasks happen at this
position.

\begin{table*}[t]
\centering
\caption{Values of the event slot \texttt{\{EVENT\}} for each task. All tasks use
Qwen3.5-2B and read layer 23. For TVSum, \texttt{\{TOPIC\}} is the title of the video.}
\label{tab:app_event}
\small
\renewcommand{\arraystretch}{1.12}
\setlength{\aboverulesep}{0pt}
\setlength{\belowrulesep}{0pt}
\begin{tabular}{p{0.2\textwidth}|p{0.74\textwidth}}
\toprule
\rowcolor{vadHeader}
\textbf{Task} & \texttt{\{EVENT\}} \\
\midrule
UCF-Crime, UBnormal, XD-Violence, UBI-Fights
& Anomalous events, restricted to six categories: violent conflict, theft and crime,
traffic accidents, personal emergencies, environmental accidents and disasters, and
prohibited public behavior. Everyday activities outside these six categories are not
anomalies. \\
TAD
& Traffic anomalies, including vehicle accidents, illegal turns, illegal lane occupation,
wrong-way driving, pedestrians on motor lanes, road debris, and other situations that
clearly obstruct safe passage. Waiting at lights, congestion, normal lane changes, legal
parking, and rainy-night or blurred frames are not anomalies. \\
TVSum
& A video highlight, i.e., a key moment that best represents the video topic
``\texttt{\{TOPIC\}}'' and is suitable for a short summary. Lead-ups, passing shots,
irrelevant frames, and repeated processes are not highlights. \\
THUMOS14
& A sports action, i.e., a person performing one of the following 20 actions: baseball
pitch, basketball dunk, billiards shot, clean and jerk, cliff diving, cricket bowling,
cricket shot, diving, frisbee catch, golf swing, hammer throw, high jump, javelin throw,
long jump, pole vault, shot put, soccer penalty, tennis swing, discus throw, volleyball
spike (definitions in \cref{tab:app_thumos}). Clips showing only spectators,
commentary, scoreboards, or empty fields, or persons waiting, walking, or celebrating
without performing these actions, are not actions. \\
\bottomrule
\end{tabular}
\end{table*}

\begin{table*}[t]
\centering
\caption{The 20 THUMOS14 actions and their definitions in \texttt{\{EVENT\}}. ID is the
THUMOS14 class index.}
\label{tab:app_thumos}
\footnotesize
\renewcommand{\arraystretch}{1.08}
\setlength{\aboverulesep}{0pt}
\setlength{\belowrulesep}{0pt}
\begin{tabular}{r|l|p{0.68\textwidth}}
\toprule
\rowcolor{vadHeader}
\textbf{ID} & \textbf{Class} & \textbf{Definition} \\
\midrule
7  & BaseballPitch & The pitcher throws the baseball toward the batter with a body turn and arm swing. \\
9  & BasketballDunk & A player jumps with the ball and slams it into the basket from above the rim. \\
12 & Billiards & A player aims and strikes a billiard ball with a cue. \\
21 & CleanAndJerk & The athlete lifts the barbell to the shoulders and then overhead; a frame may show only one phase. \\
22 & CliffDiving & The athlete jumps from a cliff or a high point over water and enters natural water. \\
23 & CricketBowling & The bowler runs up, strides, and swings the arm to deliver the ball to the batter. \\
24 & CricketShot & The batter swings or blocks the incoming ball with a flat bat. \\
26 & Diving & The athlete leaps from a pool springboard or platform, goes airborne, and enters the pool. \\
31 & FrisbeeCatch & The receiver reaches, moves, or jumps to catch a flying frisbee. \\
33 & GolfSwing & The player completes the backswing, downswing, or follow-through with a golf club. \\
36 & HammerThrow & The athlete holds the hammer handle, spins the hammer around the body, and releases it. \\
40 & HighJump & The athlete runs up, takes off, and clears the bar with the body. \\
45 & JavelinThrow & The athlete runs up with a javelin and throws it forward with a stride and arm swing. \\
51 & LongJump & The athlete runs along the runway, takes off from one foot, and lands in the sand pit. \\
68 & PoleVault & The athlete runs up with a long pole, plants it, and is lifted over the bar. \\
79 & Shotput & The athlete holds the shot near the neck and pushes it out with a leg drive, turn, and arm extension. \\
85 & SoccerPenalty & A player runs up near the penalty spot and kicks the ball at the goal. \\
92 & TennisSwing & A player swings a racket in a serve, forehand, or backhand stroke. \\
93 & ThrowDiscus & The athlete holds a disc-shaped discus and throws it with an arm swing and body turn. \\
97 & VolleyballSpiking & A player jumps and strikes the ball with an overhead arm swing toward the opponent's court. \\
\bottomrule
\end{tabular}
\end{table*}

Apart from \texttt{\{EVENT\}}, the task and question text is identical across all tasks.
All THUMOS14 classes share one \texttt{\{EVENT\}}: a single forward pass yields features
that the 20 per-class readouts of \cref{app:multiclass} share, so class identity comes from
the per-class fitting, not from the prompt. The zero-shot baseline (\cref{app:protocol})
uses the same template and takes the probability of the answer token ``yes'' as its score.

\subsection{Candidate Selection, Regularization, and Algorithms}
\label{app:algo}

\paragraph{Candidate selection.}
Neuron $i$ is scored by $\mathrm{score}_i=(\sigma_{+,i}^2/\sigma_{\mathrm{ref},i}^2)\,\sigma_{+,i}$
(\cref{sec:calibration}), and the top $|\mathcal{C}|=2048$ neurons are kept (one third of $D=6144$).
Selection uses the training set only, and $|\mathcal{C}|$ is shared by all datasets and
backbones.

\paragraph{Regularization.}
We use $\tilde S_{\mathrm{ref}}=S_{\mathrm{ref}}+\varepsilon I$ with $\varepsilon=0.25$.
Because each candidate is scaled by $\sigma_{\mathrm{ref},i}$, the diagonal of
$S_{\mathrm{ref}}$ equals 1 and $\varepsilon$ has the same relative strength in every
coordinate (\cref{app:invariance}). $\varepsilon$ is shared by all datasets and backbones.
Note that the unit normalization $\mathbf{u}_k^\top\tilde S_{\mathrm{ref}}\mathbf{u}_k=1$ holds for the
regularized matrix. The actual background variance along $\mathbf{u}_k$ is
$\mathbf{u}_k^\top S_{\mathrm{ref}}\mathbf{u}_k=1-\varepsilon\|\mathbf{u}_k\|^2<1$, so
$\lambda_k-1$ is the excess variance relative to $\tilde S_{\mathrm{ref}}$, not the exact
background excess variance (\cref{app:reg}).

\paragraph{Numerical solution.}
We eigendecompose $\tilde S_{\mathrm{ref}}$ to obtain $W=\tilde S_{\mathrm{ref}}^{-1/2}$,
eigendecompose $WS_+W$, and map back with $\mathbf{u}_k=W\mathbf{v}_k$, which gives
$\mathbf{u}_k^\top\tilde S_{\mathrm{ref}}\mathbf{u}_k=1$. Decompositions use
\texttt{torch.linalg.eigh} in float64; the online readout parameters are stored in float32.
Each direction is oriented so that the mean projection of the weak positive pool is positive.

\paragraph{Algorithm 1: offline fitting.}
\begin{enumerate}\small\itemsep0pt
\item \textbf{Input:} neuron activations $\{\mathbf{a}_t\}$ of training clips, video-level
labels $y_v$, $|\mathcal{C}|$, $\varepsilon$, $K$.
\item $\mathcal{P}_{\mathrm{ref}}\leftarrow\{t:y_v(t)=0\}$,\; $\mathcal{P}_+\leftarrow\{t:y_v(t)=1\}$.
\item Compute $\mu_{\mathrm{ref}},\sigma_{\mathrm{ref}},\mu_+,\sigma_+$ per neuron.
\item $\mathcal{C}\leftarrow$ top $|\mathcal{C}|$ neurons by $\mathrm{score}_i$.
\item $s_i\leftarrow\operatorname{sign}(\mu_{+,i}-\mu_{\mathrm{ref},i})$;\;
$x_{t,i}\leftarrow s_i(a_{t,i}-\mu_{\mathrm{ref},i})/\sigma_{\mathrm{ref},i}$, $i\in\mathcal{C}$.
\item $S_q\leftarrow|\mathcal{P}_q|^{-1}\sum_{t\in\mathcal{P}_q}\mathbf{x}_t\mathbf{x}_t^\top$,
$q\in\{\mathrm{ref},+\}$ (no mean subtraction).
\item $\tilde S_{\mathrm{ref}}\leftarrow S_{\mathrm{ref}}+\varepsilon I$;\;
$W\leftarrow\tilde S_{\mathrm{ref}}^{-1/2}$.
\item $(\lambda_k,\mathbf{v}_k)\leftarrow\mathrm{eigh}(WS_+W)$ in descending order;\;
$\mathbf{u}_k\leftarrow W\mathbf{v}_k$.
\item If $\operatorname{mean}_{t\in\mathcal{P}_+}(\mathbf{u}_k^\top\mathbf{x}_t)<0$, set
$\mathbf{u}_k\leftarrow-\mathbf{u}_k$.
\item $U_K\leftarrow[\mathbf{u}_1,\dots,\mathbf{u}_K]$,\; $w_k\leftarrow\max(\lambda_k-1,0)$.
\item \textbf{Output:} $\mathcal{C},\mu_{\mathrm{ref}},\sigma_{\mathrm{ref}},s,U_K,\mathbf{w}$.
\end{enumerate}

\paragraph{Algorithm 2: online scoring.}
\begin{enumerate}\small\itemsep0pt
\item \textbf{Input:} test clip $c_t$ and the offline parameters.
\item Run the frozen MLLM once and read $\mathbf{a}_t$ at \texttt{[Last]} (\cref{eq:neuron}).
\item $x_{t,i}\leftarrow s_i(a_{t,i}-\mu_{\mathrm{ref},i})/\sigma_{\mathrm{ref},i}$, $i\in\mathcal{C}$.
\item $\mathbf{z}_t\leftarrow U_K^\top\mathbf{x}_t$.
\item $e_t\leftarrow\|\mathrm{ReLU}(\mathbf{z}_t)\odot\sqrt{\mathbf{w}}\|_2$ (\cref{eq:energy}).
\item \textbf{Output:} $e_t$.
\end{enumerate}

\subsection{Datasets and Evaluation Protocol}
\label{app:datasets}

UCF-Crime, UBnormal, XD-Violence, UBI-Fights, TAD, and THUMOS14 use only video-level
training labels. TVSum is an extended experiment that shows NERVE applies beyond
anomalies; it is not part of the main experiments. For TVSum, the weak positive pool uses
video-level weak labels of TVSum videos, and the reference (negative) pool is built from
clips labeled non-highlight in YouTube Highlights, a separate dataset, used at the weak-label
level. Evaluation uses 5-fold cross-validation on TVSum. Temporal annotations of the test
sets are used for evaluation and visualization.

\begin{table*}[t]
\centering
\caption{Dataset settings.}
\label{tab:app_datasets}
\small
\renewcommand{\arraystretch}{1.12}
\setlength{\aboverulesep}{0pt}
\setlength{\belowrulesep}{0pt}
\begin{tabular}{p{0.11\textwidth}|p{0.17\textwidth}|p{0.24\textwidth}|p{0.2\textwidth}|p{0.15\textwidth}}
\toprule
\rowcolor{vadHeader}
\textbf{Dataset} & \textbf{Train / test split} & \textbf{Training videos (pos / neg)}
& \textbf{Label source} & \textbf{Metrics} \\
\midrule
UCF-Crime & official & 810 / 800 (1,610) & official video-level labels & frame AUC \\
UBnormal & official & 82 / 186 (268) & official & frame AUC \\
XD-Violence & official & 1,905 / 2,049 (3,954) & official & frame AP (and AUC) \\
UBI-Fights & official & 176 / 757 (933); test: 40 fight / 27 normal (67) & official & AUC / AP / BACC \\
TAD & official & 190 / 210 (400); test: 60 anomalous / 40 normal (100) & official & AUC / AP / BACC \\
TVSum & 5-fold cross-validation (evaluation) & positive pool: video-level weak labels of TVSum videos; negative pool: non-highlight clips of YouTube Highlights (weak-label level, external dataset); shot-level highlight labels binarized from official importance scores for evaluation (below) & AUC / AP / BACC \\
THUMOS14 & official: validation for training, test for testing & 200 training videos (validation), 20 classes; 213 test videos & action classes contained in each video & Frame mAP / Acc21 / Action Top-1 \\
\bottomrule
\end{tabular}
\end{table*}

\paragraph{TVSum labels.}
For each 2\,s shot, we average the importance scores (1--5) of the 20 annotators. Within each
video, the shot means are sorted in ascending order and the value at sorted index
$\lfloor n/2\rfloor$ ($n$ shots) is the threshold. Shots with mean at least the threshold
are highlights, and the others are non-highlights. These shot-level labels are used for
evaluation only. For training, the weak positive pool of NERVE uses video-level weak labels
of TVSum videos. The negative pool is built from clips labeled non-highlight in YouTube
Highlights, which is a separate dataset, and is used at the weak-label level; no TVSum
shot-level labels are used for training. Evaluation uses 5-fold cross-validation on TVSum, with
10 test videos in each fold, and we report the mean over the folds
(\cref{tab:additional_video}). TVSum serves as an extended demonstration of generality and
not as a main experiment.

\paragraph{THUMOS14 frame-level scores.}
The score of a clip is assigned to the 16 frames it covers.

\paragraph{Metrics.}
\begin{itemize}\itemsep0pt
\item \textbf{AUC / AP.} On TVSum, shot scores and labels of all test videos in a fold are
concatenated before computing the metric, and the five folds are averaged.
\item \textbf{BACC.} We sweep the decision threshold on the test set and report the
maximum balanced accuracy.
\item \textbf{Frame mAP.} The mean of the per-class frame-level AP over 20 actions.
\item \textbf{Acc21.} Clip-level accuracy over 21 classes (20 actions plus background).
\item \textbf{Action Top-1.} Computed on clips whose ground truth is an action; a
prediction is correct if its class is any ground-truth action class of the clip.
\end{itemize}

\subsection{Multi-Class Extension (THUMOS14)}
\label{app:multiclass}

For each class $j$, the weak positive pool consists of clips from training videos that
contain class $j$, and the reference pool of clips from the remaining training videos.
Algorithm~1 is run per class to obtain $U_K^{(j)}$ and $\mathbf{w}^{(j)}$, and online the
per-class projections are stacked into one matrix multiplication that yields $e_t^{(j)}$.
\begin{itemize}\itemsep0pt
\item \textbf{Clip-level class.} For an action clip, $\hat y_t=\operatorname{arg\,max}_j
e_t^{(j)}$. Actions and background are separated by a global action score. The
threshold is chosen as the one that maximizes the binary action/background accuracy on the
test set.
\item \textbf{Frame-level evaluation.} Class scores of a clip are assigned to its frames,
the frame-level AP is computed per class, and Frame mAP is the mean over 20 classes.
\item \textbf{Video-level class and temporal intervals.} Video-level classes are obtained
by aggregating $e_t^{(j)}$ within a video, and intervals by merging consecutive
high-energy clips. Both only aggregate and group energies, without gradient updates.
The Frame mAP, Acc21, and Action Top-1 in \cref{tab:additional_video} are computed at the
clip and frame level and do not depend on these two steps.
\end{itemize}

\subsection{Comparison Protocol}
\label{app:protocol}

\paragraph{Zero-shot baseline.}
The baseline uses the same frozen backbone, input, and template as NERVE (\cref{app:template})
and scores a clip by the probability $P(\text{yes})$ of the answer token.

\paragraph{Tuning-free.}
In \cref{tab:vad_comparison}, tuning-free means no task-specific gradient updates to any
model component. NERVE satisfies this and uses video-level training labels only to fit
closed-form statistics, whereas methods such as LLaVA-1.5 and EventVAD use no training labels.

\paragraph{Baseline numbers.}
All baseline numbers in \cref{tab:vad_comparison} are taken from the original papers.
$^\dagger$ marks the best configuration selected per dataset in the original paper (PRISM,
URF), and $^{*}$ marks a number reported by LRPO with guiding questions transferred from
XD-Violence.

\subsection{Efficiency Measurement}
\label{app:efficiency}

The measurements of \cref{sec:exp_eff} and \cref{tab:efficiency} are on XD-Violence.
\begin{itemize}\itemsep0pt
\item \textbf{Hardware and setting.} One A100 80GB GPU, batch size 1, frozen Qwen3.5-2B.
We run 10 warm-up iterations and average the latency over 200 clips.
\item \textbf{AR caption baseline.} The same backbone and input, greedy decoding of
64 tokens, excluding downstream LLM scoring. It is therefore a lower bound on the latency
of generative methods (2062.7\,ms).
\item \textbf{Latency breakdown.} Preprocessing (13.3\,ms) + backbone forward (231.7\,ms) +
readout (0.57\,ms) = 245.6\,ms. The readout covers candidate extraction, calibration,
projection, and energy computation.
\item \textbf{RTF.} Clip duration over per-clip latency: 1\,s / 245.6\,ms = $4.07\times$.
\item \textbf{TTFS.} The time from the start of the stream to the first score, including
1\,s of clip buffering, measured end to end (NERVE 1.27\,s, AR 3.46\,s). It exceeds
``1\,s + steady-state latency'' (NERVE 1.25\,s, AR 3.06\,s) by 0.02\,s and 0.40\,s, which
is the first-call overhead of the first clip.
\item \textbf{Peak memory.} Recorded with \texttt{torch.cuda.max\_memory\_allocated}:
4.28\,GB for NERVE and 4.30\,GB for the AR baseline, so the readout adds no memory.
\item \textbf{Offline fitting.} Algorithm~1 on the cached activations of 682,773 training
clips takes 16.9\,s with 2.87\,GB peak memory, excluding the forward passes for feature
extraction.
\item \textbf{Readout parameters (0.28\,MB).} $\mu_{\mathrm{ref}},\sigma_{\mathrm{ref}}$ ($|\mathcal{C}|$-dim each),
$U_K$ ($|\mathcal{C}|\times K$), and $\mathbf{w}$ ($K$-dim) in float32, and $s$ in int8.
With $|\mathcal{C}|=2048$ and $K=32$ this is $4\,(2\cdot2048+2048\cdot32+32)+2048=280{,}704$
bytes, about 0.28\,MB.
\end{itemize}

\subsection{Reproducibility}
\label{app:repro}

The fit of NERVE is closed-form with no random initialization or iterative optimization,
so the result is deterministic given the cached features. The only randomness is in the
controlled-dilution resampling of \cref{fig:recovery} and the label permutations of
\cref{tab:eigenspectrum}, which use a fixed seed of 42. Code and configurations will be
released upon acceptance.

\section{Choice of Readout Source}
\label{app:source}

We compare three readout sources under identical pools and readout algorithm
(\cref{sec:exp_mllm}). All are taken from the last decoder layer at the \texttt{[Last]} position:
\begin{itemize}\itemsep0pt
\item \textbf{MLP neuron (NERVE):} the activation $\mathbf{a}_t\in\mathbb{R}^{D}$ after the
gated product and before the down-projection;
\item \textbf{Residual hidden state:} the residual-stream hidden state output by the layer,
of dimension $d$;
\item \textbf{Attention output:} the output of the attention module of the layer, of
dimension $d$.
\end{itemize}
The three sources share the training pools, calibration, candidate selection rule,
candidate set size $|\mathcal{C}|$, $K=32$, and Algorithms~1--2. Only the source differs.

\begin{table}[!ht]
\centering
\caption{Readout sources on three backbones (\%, frame level). The MLP neuron rows are the
NERVE results in the main paper.}
\label{tab:app_source}
\setlength{\tabcolsep}{3.5pt}
\renewcommand{\arraystretch}{1.12}
\setlength{\aboverulesep}{0pt}
\setlength{\belowrulesep}{0pt}
\resizebox{\columnwidth}{!}{%
\begin{tabular}{l|l|cc|c|c}
\toprule
\rowcolor{vadHeader}
& & \multicolumn{2}{c|}{\textbf{XD-Violence}} & \textbf{UCF-Crime} & \textbf{UBnormal} \\
\rowcolor{vadHeader}
\textbf{Backbone} & \textbf{Source} & AUC & AP & AUC & AUC \\
\midrule
& Attention output & 92.84 & 75.62 & 85.19 & 70.43 \\
& Residual hidden state & 93.58 & 77.87 & 85.84 & 71.12 \\
\multirow{-3}{*}{Qwen3.5-2B}
& \cellcolor{vadOurs}MLP neuron (NERVE) & \cellcolor{vadOurs}\textbf{95.26}
& \cellcolor{vadOurs}\textbf{84.24} & \cellcolor{vadOurs}\textbf{87.53}
& \cellcolor{vadOurs}\textbf{72.78} \\
\midrule
& Attention output & 93.12 & 76.45 & 83.74 & 70.85 \\
& Residual hidden state & 93.41 & 78.19 & 84.62 & 71.38 \\
\multirow{-3}{*}{InternVL3.5-2B}
& \cellcolor{vadOurs}MLP neuron (NERVE) & \cellcolor{vadOurs}\textbf{95.25}
& \cellcolor{vadOurs}\textbf{84.90} & \cellcolor{vadOurs}\textbf{86.16}
& \cellcolor{vadOurs}\textbf{72.96} \\
\midrule
& Attention output & 88.06 & 64.18 & 76.85 & 66.12 \\
& Residual hidden state & 88.79 & 65.94 & 77.63 & 66.95 \\
\multirow{-3}{*}{Gemma4-E2B}
& \cellcolor{vadOurs}MLP neuron (NERVE) & \cellcolor{vadOurs}\textbf{90.23}
& \cellcolor{vadOurs}\textbf{70.39} & \cellcolor{vadOurs}\textbf{79.92}
& \cellcolor{vadOurs}\textbf{68.67} \\
\bottomrule
\end{tabular}%
}
\end{table}

On all three backbones and all four metrics, MLP neurons outperform the other two sources
(\cref{tab:app_source}). Relative to the second-best source, the residual hidden state,
MLP neurons raise AUC by 1.44--2.29 points and XD-Violence AP by 4.45--6.71 points. The
residual hidden state in turn outperforms the attention output in all 12 comparisons.

\section{Ablations and Sensitivity}
\label{app:ablation}

All experiments in this section use cached neuron features of Qwen3.5-2B. Each closed-form
refit takes only seconds.

\subsection{Component Ablation}
\label{app:components}

Each variant changes one component and keeps the rest of NERVE (\cref{tab:app_ablation}).

\begin{table}[!ht]
\centering
\caption{Component ablation (\%, frame level). Random candidates are the mean $\pm$ std
over 5 seeds.}
\label{tab:app_ablation}
\setlength{\tabcolsep}{3pt}
\renewcommand{\arraystretch}{1.12}
\setlength{\aboverulesep}{0pt}
\setlength{\belowrulesep}{0pt}
\resizebox{\columnwidth}{!}{%
\begin{tabular}{l|l|cc|c|c}
\toprule
\rowcolor{vadHeader}
& & \multicolumn{2}{c|}{\textbf{XD-Violence}} & \textbf{UCF} & \textbf{UBnormal} \\
\rowcolor{vadHeader}
\textbf{Variant} & \textbf{Change} & AUC & AP & AUC & AUC \\
\midrule
Per-pool centering & each pool subtracts its own mean & 95.05 & 83.44 & 85.62 & 69.79 \\
No $\sigma_{\mathrm{ref}}$ scaling & no division by $\sigma_{\mathrm{ref},i}$ in \cref{eq:calib} & 94.02 & 76.34 & 86.47 & 69.60 \\
No whitening & PCA directly on $S_+$ & 92.34 & 71.73 & 86.13 & 69.50 \\
Equal weights & $w_k\equiv1$ & 94.26 & 79.85 & 87.11 & 70.62 \\
Two-sided energy & no ReLU in \cref{eq:energy} & 94.85 & 82.91 & 86.39 & 71.54 \\
Random candidates & $|\mathcal{C}|$ random neurons & 94.92{\scriptsize$\pm$0.19} & 82.83{\scriptsize$\pm$0.81} & 86.90{\scriptsize$\pm$0.09} & 70.82{\scriptsize$\pm$0.56} \\
\midrule
\rowcolor{vadOurs}
\textbf{NERVE} & -- & \textbf{95.26} & \textbf{84.24} & \textbf{87.53} & \textbf{72.78} \\
\bottomrule
\end{tabular}%
}
\end{table}

The full method is the best on all four metrics. Removing whitening causes the largest drop
(XD-Violence AP $-12.51$), followed by removing $\sigma_{\mathrm{ref}}$ scaling ($-7.90$) and
equal weights ($-4.39$). Per-pool centering is 1.91 and 2.99 points lower on UCF-Crime and
UBnormal, consistent with the rotation of directions analyzed in \cref{app:centering}.
The $\sigma_{\mathrm{ref}}$ scaling also sets the relative strength of the regularizer in
each coordinate (\cref{app:invariance}), so that row reflects both effects.

\subsection{Alternative Weak-Label Readouts}
\label{app:alternatives}

We compare NERVE with readouts one could use instead, on the same frozen Qwen3.5-2B
candidate features $\mathbf{x}_t$ and the same video-level labels:
\begin{itemize}\itemsep0pt
\item \textbf{Mean-difference direction:} $\mathbf{u}=\bar{\mathbf{x}}_+-\bar{\mathbf{x}}_{\mathrm{ref}}$,
first-order statistics only;
\item \textbf{Broadcast linear probe:} each clip inherits its video label, and a logistic
regression is trained;
\item \textbf{MIL top-$k$:} a linear head whose video score is the mean of the top-$k$
clip scores, trained with video labels;
\item \textbf{Per-pool centering GED:} the first row of \cref{tab:app_ablation}.
\end{itemize}

\begin{table}[!ht]
\centering
\caption{Alternative weak-label readouts (\%, frame level). Solve time covers only the
solving step (eigendecomposition or iterative training), not the accumulation of statistics
from cached activations.}
\label{tab:app_alternatives}
\setlength{\tabcolsep}{3pt}
\renewcommand{\arraystretch}{1.12}
\setlength{\aboverulesep}{0pt}
\setlength{\belowrulesep}{0pt}
\resizebox{\columnwidth}{!}{%
\begin{tabular}{l|c|cc|c|c|c}
\toprule
\rowcolor{vadHeader}
& & \multicolumn{2}{c|}{\textbf{XD-Violence}} & \textbf{UCF} & \textbf{UBnormal} & \textbf{Solve} \\
\rowcolor{vadHeader}
\textbf{Method} & \textbf{Iterative} & AUC & AP & AUC & AUC & \textbf{(s)} \\
\midrule
Mean-difference direction & no & 91.82 & 71.94 & 85.92 & 70.37 & 0.3 \\
Broadcast linear probe & yes & 94.15 & 80.56 & 86.21 & 71.90 & 4.4 \\
MIL top-$k$ & yes & 92.46 & 72.54 & 86.73 & 71.84 & 3.8 \\
Per-pool centering GED & no & 95.05 & 83.44 & 85.62 & 69.79 & 0.6 \\
\rowcolor{vadOurs}
\textbf{NERVE} & no & \textbf{95.26} & \textbf{84.24} & \textbf{87.53} & \textbf{72.78} & 0.6 \\
\bottomrule
\end{tabular}%
}
\end{table}

NERVE is the best on all four metrics without iterative training
(\cref{tab:app_alternatives}). Among the trained readouts, the broadcast linear probe is
closest on XD-Violence (AP 3.68 lower), while MIL top-$k$ reaches 72.54 AP, 11.70 below
NERVE. The first-order mean-difference direction is the weakest. Per-pool centering GED is
also closed-form and close on XD-Violence (AP 0.80 lower), but 1.91 and 2.99 points lower on
UCF-Crime and UBnormal. The 0.6\,s is the solving step only; the full offline fit including
statistics accumulation takes 16.9\,s (\cref{app:efficiency}).

\subsection{Hyperparameter Sensitivity}
\label{app:hyper}

The sensitivity to $K$ is in \cref{fig:sensK}. Here we vary the candidate set size
$|\mathcal{C}|$ and the regularization strength $\varepsilon$, both of which only need a
refit on cached features (\cref{tab:app_hyper}).

\begin{table}[!ht]
\centering
\caption{Sensitivity to $|\mathcal{C}|$ and $\varepsilon$ (XD-Violence AP, \%, frame level).
Bold marks the defaults $|\mathcal{C}|=2048$ and $\varepsilon=0.25$.}
\label{tab:app_hyper}
\setlength{\tabcolsep}{4pt}
\renewcommand{\arraystretch}{1.12}
\setlength{\aboverulesep}{0pt}
\setlength{\belowrulesep}{0pt}
\resizebox{\columnwidth}{!}{%
\begin{tabular}{l|cccccc}
\toprule
\rowcolor{vadHeader}
$|\mathcal{C}|$ & 256 & 512 & 1024 & \textbf{2048} & 4096 & 6144 \\
\midrule
XD AP & 82.85 & 84.27 & 83.87 & \textbf{84.24} & 83.70 & 83.36 \\
\midrule
\rowcolor{vadHeader}
$\varepsilon$ & 0.05 & 0.1 & \textbf{0.25} & 0.5 & 1 & 2 \\
\midrule
XD AP & 83.72 & 83.96 & \textbf{84.24} & 83.28 & 84.75 & 82.81 \\
\bottomrule
\end{tabular}%
}
\end{table}

As $|\mathcal{C}|$ goes from 256 to 6144 (all neurons for $D=6144$), XD-Violence AP stays
within 82.85--84.27. As $\varepsilon$ goes from 0.05 to 2, it stays within 82.81--84.75.
Both ranges are under 2 points, so NERVE is insensitive to these two settings. The defaults
are fixed values and are not tuned on the test set.

\section{Theoretical Details}
\label{app:theory}

This section supports \cref{sec:finding2} and \cref{sec:calibration}. It gives a
population-level derivation together with a heuristic finite-sample discussion.

\subsection{Setting and Assumptions}
\label{app:assumptions}

Let $\mathbf{x}\in\mathbb{R}^{|\mathcal{C}|}$ be the calibrated feature. By \cref{eq:calib}
the reference mean is zero, so the second moment about the reference mean is
$S_q=\mathbb{E}_q[\mathbf{x}\mathbf{x}^\top]$. Let $\pi\in(0,1]$ be the event fraction of
the weak positive pool.
\begin{itemize}\itemsep0pt
\item \textbf{(A1) Distribution consistency.} Non-event clips in the weak positive pool
follow the reference distribution, with second moment $S_{\mathrm{ref}}$.
\item \textbf{(A2) Shared coordinates.} Both pools take second moments about the
reference mean and are not centered on their own means.
\item \textbf{(A3) Finite samples.} $S_{\mathrm{ref}}$ and $S_+$ are estimated from
finitely many clips, and $S_{\mathrm{ref}}$ is regularized to $\tilde S_{\mathrm{ref}}$.
\end{itemize}
\cref{app:dilution,app:centering,app:invariance} assume (A1)(A2), exact moments, and no
regularization ($\varepsilon=0$), so the results there are exact only in that setting.
\cref{app:reg,app:finite} discuss (A3), including how regularization changes the dilution
law, and \cref{app:shift} covers the case where (A1) fails.

\subsection{Dilution Rescales but Does Not Rotate}
\label{app:dilution}

Let $S_{\mathrm{event}}=\mathbb{E}_{\mathrm{event}}[\mathbf{x}\mathbf{x}^\top]
=\Sigma_{\mathrm{event}}+\boldsymbol\delta\boldsymbol\delta^\top$, where $\boldsymbol\delta$
is the event mean (relative to the reference mean) and $\Sigma_{\mathrm{event}}$ the event
covariance. Since $S_{\mathrm{event}}$ is a second moment about the reference mean, it
includes the mean shift $\boldsymbol\delta$.

\textbf{Proposition 1.} Under (A1)(A2), $S_+=(1-\pi)S_{\mathrm{ref}}+\pi S_{\mathrm{event}}$
(\cref{eq:mixture}). Let $W=S_{\mathrm{ref}}^{-1/2}$ and let
$M=WS_{\mathrm{event}}W$ have eigenpairs $(\lambda_k(1),\mathbf{v}_k)$. Then $WS_+W$ has
eigenvectors $\mathbf{v}_k$ and eigenvalues $\lambda_k(\pi)=1+\pi(\lambda_k(1)-1)$
(\cref{eq:linear_law}).

\textit{Proof.} The weak positive pool mixes background and event clips, so its second
moment is the proportional combination of the two, and (A1) gives the mixture form. After
whitening, $WS_+W=(1-\pi)I+\pi M$. For any eigenvector $\mathbf{v}_k$ of $M$,
$WS_+W\mathbf{v}_k=[(1-\pi)+\pi\lambda_k(1)]\mathbf{v}_k$. The identity does not change
eigenvectors, so they do not depend on $\pi$. For $\pi>0$ the map
$\lambda_k(1)\mapsto\lambda_k(\pi)$ is strictly increasing, so the eigenvalue order is
preserved and the top-$K$ subspace equals the oracle ($\pi=1$) subspace. In the original
coordinates the generalized eigenvectors are $\mathbf{u}_k=W\mathbf{v}_k$. $\square$

\subsection{Why Per-Pool Centering Rotates the Directions}
\label{app:centering}

If the weak positive pool subtracts its own mean $\boldsymbol\mu_+=\pi\boldsymbol\delta$,
its covariance is
\begin{equation}
\bar S_+=S_+-\boldsymbol\mu_+\boldsymbol\mu_+^\top
=(1-\pi)S_{\mathrm{ref}}+\pi\Sigma_{\mathrm{event}}+\pi(1-\pi)\,\boldsymbol\delta\boldsymbol\delta^\top .
\end{equation}
After whitening it becomes
$(1-\pi)I+\pi\big[W\Sigma_{\mathrm{event}}W+(1-\pi)\,W\boldsymbol\delta\boldsymbol\delta^\top W\big]$.
The relative weight $(1-\pi)$ of the rank-one term in the bracket depends on $\pi$, so the
eigenvectors rotate with $\pi$. As $\pi\to0$ the bracket tends to $WS_{\mathrm{event}}W$,
whose directions are those of NERVE. As $\pi$ grows, the weight of the mean-shift direction
falls, and at $\pi\to1$ only the event covariance remains. The directions stay fixed only
if $\boldsymbol\delta=0$ or $W\boldsymbol\delta$ is an eigenvector of
$W\Sigma_{\mathrm{event}}W$. The first row of \cref{tab:app_ablation} gives the
corresponding empirical result.

\subsection{Invariance to the Calibration Transform}
\label{app:invariance}

The scaling and polarity alignment in \cref{eq:calib} form an invertible diagonal transform
$\mathbf{x}'=D\mathbf{x}$ with $D=\operatorname{diag}(s_i/\sigma_{\mathrm{ref},i})$. Then
$S'_q=DS_qD$, and the problem $DS_+D\mathbf{u}'=\lambda DS_{\mathrm{ref}}D\mathbf{u}'$ is
equivalent to $S_+(D\mathbf{u}')=\lambda S_{\mathrm{ref}}(D\mathbf{u}')$. The eigenvalues are
unchanged, the eigenvectors satisfy $\mathbf{u}'=D^{-1}\mathbf{u}$, and the projection
$\mathbf{u}'^\top\mathbf{x}'=\mathbf{u}^\top\mathbf{x}$ is unchanged.

With regularization the invariance no longer holds in general:
$D(S_{\mathrm{ref}}+\varepsilon I)D=DS_{\mathrm{ref}}D+\varepsilon D^2\neq DS_{\mathrm{ref}}D+\varepsilon I$
unless $D^2\propto I$. Scaling therefore sets the relative strength of the regularizer in each
coordinate. After scaling by $\sigma_{\mathrm{ref},i}$, every coordinate has unit variance and
$\varepsilon I$ treats all neurons alike, which is why scaling keeps large-scale neurons from
dominating the regularization (\cref{sec:calibration}).

\subsection{Effect of Regularization}
\label{app:reg}

Let $\tilde S_{\mathrm{ref}}=S_{\mathrm{ref}}+\varepsilon I$ and
$\lambda_{\min}=\lambda_{\min}(S_{\mathrm{ref}})>0$. From
$S_{\mathrm{ref}}\preceq\tilde S_{\mathrm{ref}}\preceq(1+\varepsilon/\lambda_{\min})S_{\mathrm{ref}}$
and the Courant--Fischer min-max theorem, the generalized eigenvalues satisfy
\begin{equation}
\frac{\lambda_k}{1+\varepsilon/\lambda_{\min}}\le\tilde\lambda_k\le\lambda_k ,
\end{equation}
so regularization shrinks all eigenvalues by at most a factor $1+\varepsilon/\lambda_{\min}$.
Eigenvalues along directions in which $S_{\mathrm{ref}}$ has large variance change little,
while near-singular directions are suppressed. The latter are the least reliable under finite
samples, so regularization trades a bias for lower variance. This statement is about
eigenvalues only; the effect on the eigenvectors is discussed next.

\paragraph{Background variance.}
Since $\mathbf{u}_k^\top\tilde S_{\mathrm{ref}}\mathbf{u}_k=1$, the background variance along
$\mathbf{u}_k$ is $\mathbf{u}_k^\top S_{\mathrm{ref}}\mathbf{u}_k=1-\varepsilon\|\mathbf{u}_k\|^2<1$.
It equals 1 only for $\varepsilon=0$. Hence $\lambda_k-1$ is not the exact background excess
variance, and a Gaussian approximation with unit background variance is only approximate.

\paragraph{Dilution law under regularization.}
With $W=(S_{\mathrm{ref}}+\varepsilon I)^{-1/2}$ and $M=WS_{\mathrm{event}}W$,
\begin{equation}
WS_+W=(1-\pi)\big(I-\varepsilon W^2\big)+\pi M .
\end{equation}
For $\varepsilon=0$ the first term is $(1-\pi)I$, and \cref{eq:linear_law} and the
$\pi$-invariance of the eigenvectors are exact. For $\varepsilon>0$, $I-\varepsilon W^2$ is not
a scalar matrix, so the eigenvectors in general change with $\pi$ and \cref{eq:linear_law}
holds only approximately. Let $\delta_\varepsilon$ be the spread of the eigenvalues of
$\varepsilon W^2$, which is at most $\varepsilon/(\lambda_{\min}(S_{\mathrm{ref}})+\varepsilon)$,
and let $g$ be the $K$-th eigengap of $M$. By the Davis--Kahan theorem,
$\|\sin\Theta\|\lesssim(1-\pi)\,\delta_\varepsilon/(\pi g)$, which is looser at small $\pi$ and
large $\varepsilon$. We do not quantify this approximation analytically. The sensitivity
to $\varepsilon$ in \cref{app:hyper} is only indirect evidence, since it measures
detection performance and not direction error.

\subsection{Finite Samples}
\label{app:finite}

\paragraph{Eigengap.}
By \cref{eq:linear_law}, the gap between the $K$-th and $(K{+}1)$-th whitened eigenvalues is
\begin{equation}
\mathrm{gap}_K(\pi)=\pi\big(\lambda_K(1)-\lambda_{K+1}(1)\big),
\end{equation}
which shrinks linearly with $\pi$. Let $E=\widehat{WS_+W}-WS_+W$ be the estimation error.
By the Davis--Kahan $\sin\Theta$ theorem, the principal angles between the estimated and
true subspaces satisfy $\|\sin\Theta\|\lesssim\|E\|/\mathrm{gap}_K(\pi)$. For a fixed
sample size, $\|E\|$ is of order $O(\sqrt{|\mathcal{C}|/n_+})$, so the bound grows as
$1/\pi$. This matches \cref{fig:recovery}, where subspace similarity decreases and its
variance grows at small $\pi$.

\paragraph{Selection bias.}
The candidate set $\mathcal{C}$ is chosen by variance ratio on the same training data,
which inflates $\hat\lambda_k$. The permutation null of \cref{sec:finding1} permutes
video-level labels and \emph{reruns the full pipeline including selection}, so the null
distribution contains the same selection bias. The ratio of $\lambda_1$ to the permutation
maximum in \cref{tab:eigenspectrum} (22--42$\times$) therefore already accounts for it.

\subsection{Definitions of Quantities in the Main Text}
\label{app:defs}

\begin{itemize}\itemsep0pt
\item \textbf{Subspace similarity} (\cref{fig:recovery}, right). In the whitened coordinates
of the oracle, let $U,V\in\mathbb{R}^{|\mathcal{C}|\times K}$ be the orthonormalized top-$K$
directions of the weak-label fit and of the oracle. The similarity is
$\|U^\top V\|_F^2/K\in[0,1]$, the mean squared cosine of the principal angles, with $K=4$.
\item \textbf{PR.} \cref{eq:pr}, $\mathrm{PR}=(\sum_kw_k)^2/\sum_kw_k^2$ with
$w_k=\max(\lambda_k-1,0)$.
\item \textbf{Permutation maximum.} Video-level training labels are permuted 20 times, the
full pipeline is rerun each time, and the maximum $\lambda_1$ over the 20 runs is taken.
\item \textbf{Linear-law deviation (6\%).} For $\pi\ge0.3$, the relative deviation
$|\hat\lambda_1(\pi)-[1+\pi(\hat\lambda_1(1)-1)]|/[1+\pi(\hat\lambda_1(1)-1)]$ is within 6\%.
\item \textbf{86.5\%.} The whitened background on $(\mathbf{u}_1,\mathbf{u}_2)$ is
approximately a 2D standard normal, so $r^2=z_1^2+z_2^2\sim\chi^2_2$ and
$P(r\le2)=1-e^{-2}\approx86.5\%$. The measured value is 89.6\% (\cref{sec:finding1}).
\item \textbf{Within-video AUC 80.7\%.} On the 462 test videos that contain both event and
background clips, the score is the mean activation of the 32 neurons with the highest
loading on $\mathbf{u}_1$; the clip-level AUC is computed per video and averaged
(\cref{sec:finding1}).
\end{itemize}

\subsection{When Distribution Consistency Fails}
\label{app:shift}

Suppose the background in the weak positive pool has second moment $S_{\mathrm{ref}}+\Delta$
(for example, positive videos come from systematically different scenes). Then
\begin{equation}
WS_+W=I+(1-\pi)\,W\Delta W+\pi\,(M-I).
\end{equation}
The leading directions of $W\Delta W$ enter the readout with weight $(1-\pi)$, so the
effect is larger at small $\pi$, and scene directions can mix into the event subspace.
A local analogue appears at test time: a background state that is rare in the reference
pool (strong camera motion, a scene cut) gives a large $\|W\mathbf{x}_t\|$ and may project
onto $+\mathbf{u}_k$, producing the false peak in the last panel of \cref{fig:qual}. This
links the failure case of \cref{sec:exp_qual} to condition (i) of \cref{sec:finding2}.

\section{Visualizations and Neuron-Space Geometry}
\label{app:vis}

\subsection{More Neuron Activation Surfaces}
\label{app:surface}

\begin{figure*}[t]
  \centering
  \includegraphics[width=0.95\linewidth]{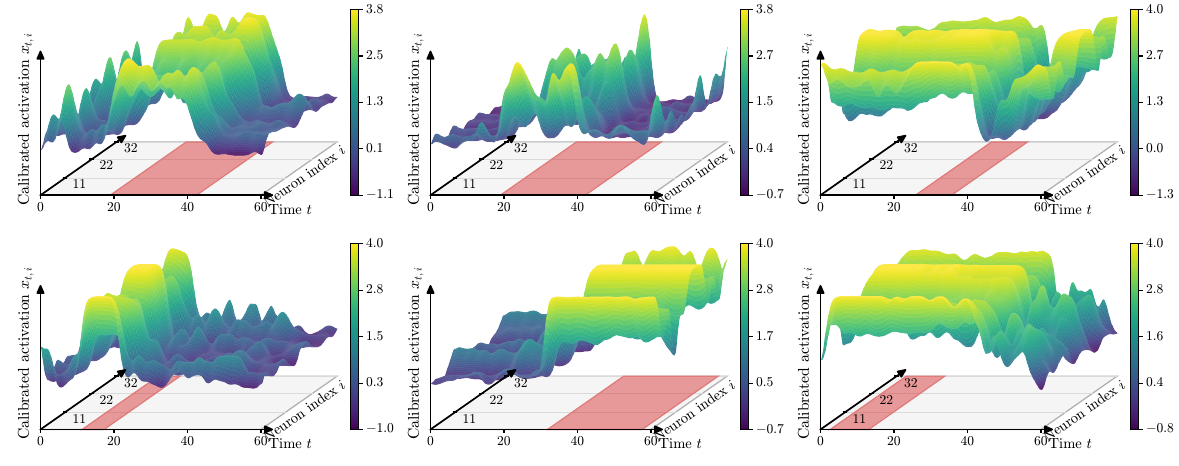}
  \caption{\textbf{More examples of event response patterns.} Calibrated activations of the
  top-32 neurons ranked by loading on $\mathbf{u}_1$ for six additional test videos
  (Qwen3.5-2B, layer 23). Red floor: ground-truth event.}
  \label{fig:app_surface}
\end{figure*}

\cref{fig:app_surface} shows more test videos in the format of \cref{fig:findings}(a).
Neurons are sorted by their loading $|u_{1,i}|$ on $\mathbf{u}_1$ in descending order, and
the top 32 are shown. The red floor marks the ground-truth event. In each panel, the top-loading neurons change together at the
boundary of the event.

\subsection{More Event Score Curves}
\label{app:curves}

\begin{figure*}[t]
  \centering
  \includegraphics[width=0.74\linewidth]{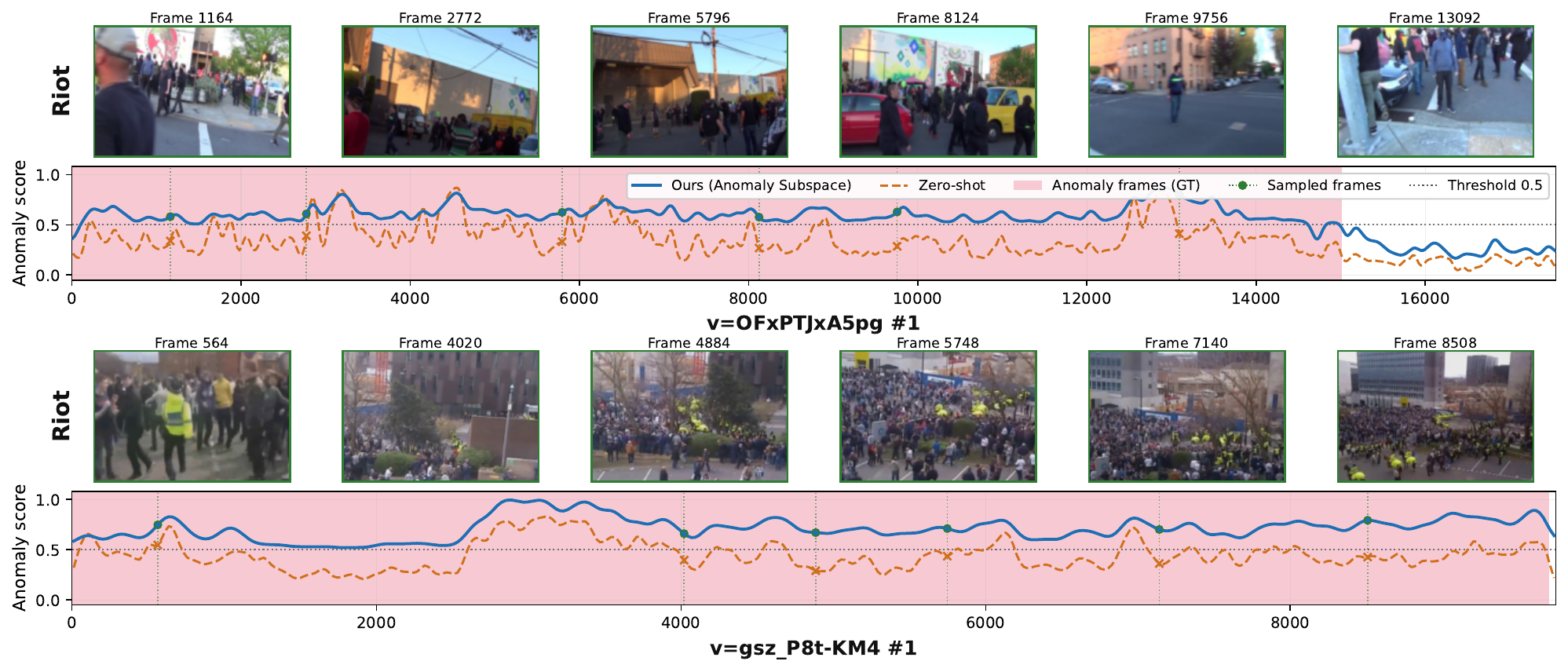}\\[2pt]
  \includegraphics[width=0.74\linewidth]{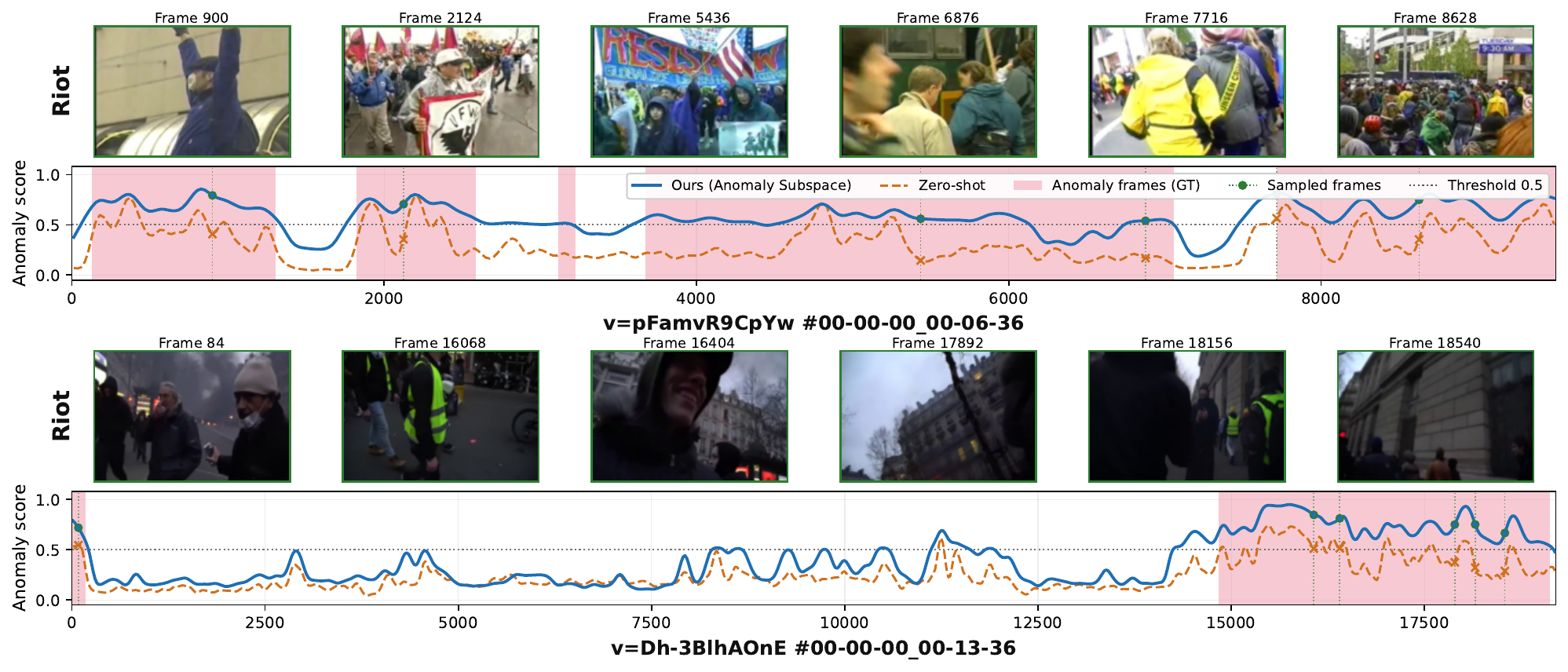}
  \caption{\textbf{Event score curves on XD-Violence.} Four test videos of the
  \emph{Riot} class. Shaded: ground-truth anomalous frames; blue: NERVE; orange dashed:
  zero-shot $P(\text{yes})$; dotted: threshold 0.5.}
  \label{fig:app_xd}
\end{figure*}

\begin{figure*}[t]
  \centering
  \includegraphics[width=\linewidth]{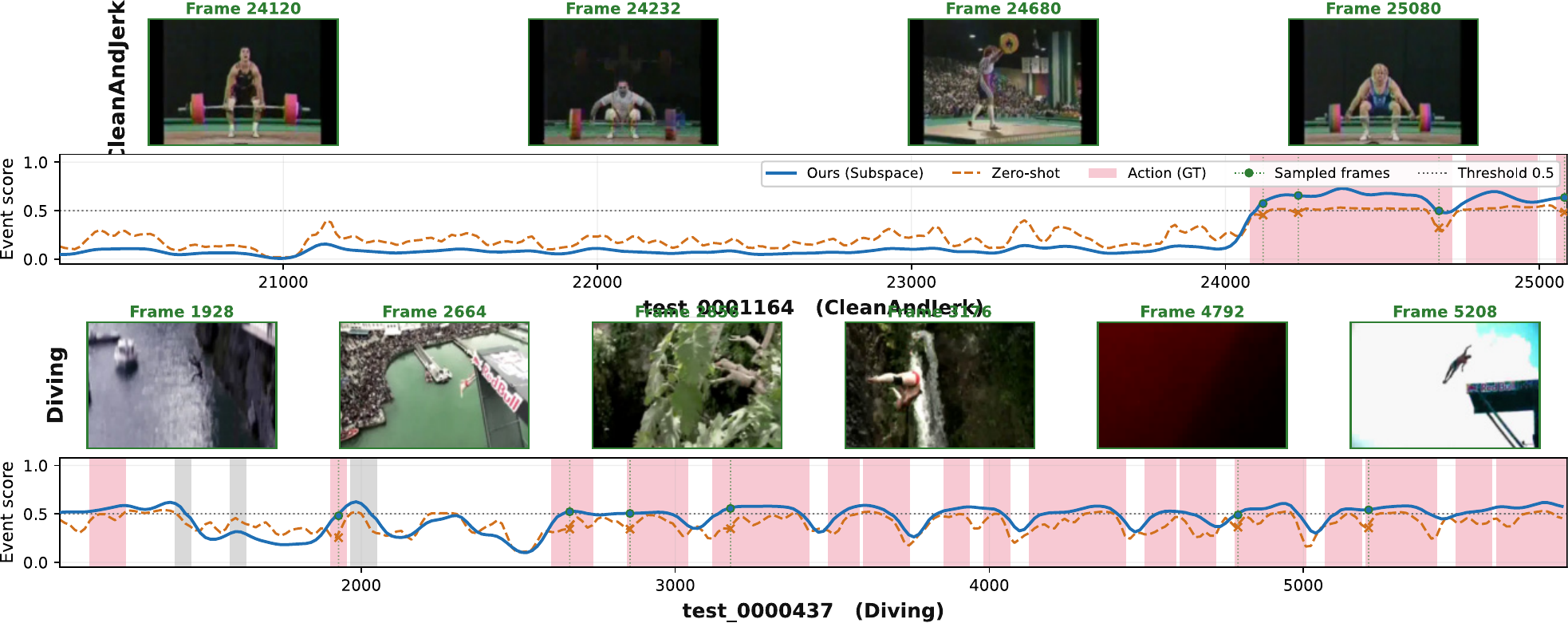}
  \caption{\textbf{Event score curves on THUMOS14.} Test videos of the \emph{CleanAndJerk}
  (\texttt{test\_0001164}) and \emph{Diving} (\texttt{test\_0000437}) classes. Shaded:
  ground-truth action intervals; blue: NERVE; orange dashed: zero-shot $P(\text{yes})$;
  dotted: threshold 0.5.}
  \label{fig:app_thumos}
\end{figure*}

\cref{fig:app_xd} shows the event score $e_t$ on further XD-Violence test videos,
and \cref{fig:app_thumos} on THUMOS14. In each figure, the frames above the curves are
sampled from the video and shaded regions mark ground-truth intervals, used only for
visualization. The blue curve is $e_t$ of NERVE and the orange dashed curve is the zero-shot
$P(\text{yes})$ of the same backbone; the dotted line is the threshold 0.5. The curves
include events that occupy only a small part of a long video (low $\pi$).

\subsection{Geometry of the Neuron Space}
\label{app:geometry}

\begin{figure*}[t]
  \centering
  \includegraphics[width=\linewidth]{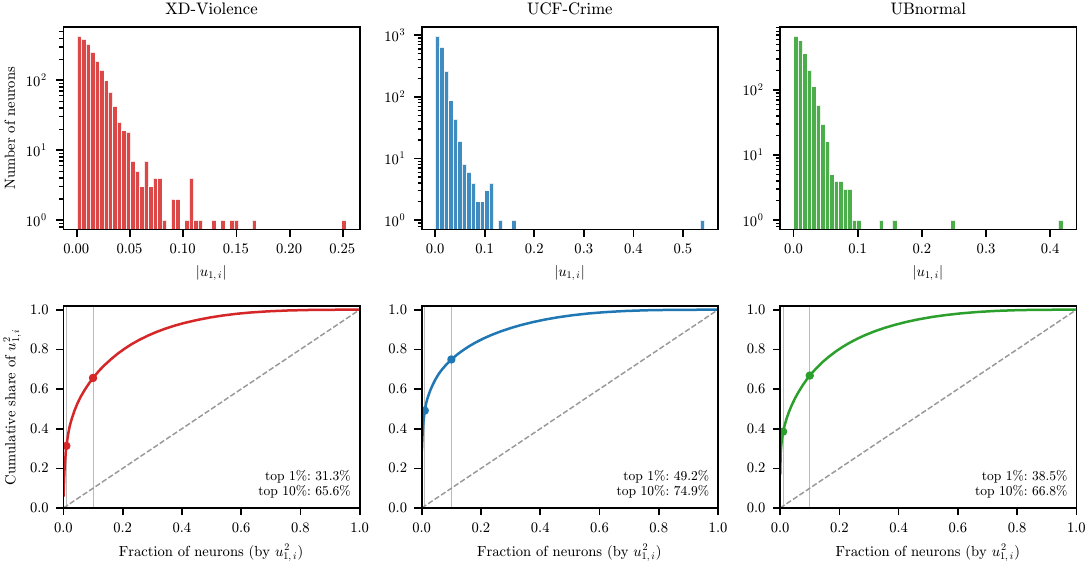}
  \caption{\textbf{Loading distribution of $\mathbf{u}_1$} on XD-Violence, UCF-Crime, and
  UBnormal (Qwen3.5-2B). Top: histogram of $|u_{1,i}|$ (log count). Bottom: cumulative share
  of the squared loading against the fraction of neurons ranked by $u_{1,i}^2$; the dashed
  diagonal is a uniform loading.}
  \label{fig:app_geometry}
\end{figure*}

The following statistics use the fitted $U_K$ of Qwen3.5-2B and need no further experiments.
\begin{itemize}\itemsep0pt
\item \textbf{Sparsity of loadings.} \cref{fig:app_geometry} shows the histogram of
$|u_{1,i}|$ (top) and the cumulative share of the squared loading against the fraction of
neurons ranked by $u_{1,i}^2$ (bottom) on XD-Violence, UCF-Crime, and UBnormal. Most neurons
have small loadings, and a small fraction of neurons carries a large share of the squared
loading: the top 1\% of neurons holds 31.3\%, 49.2\%, and 38.5\% of the squared loading on
XD-Violence, UCF-Crime, and UBnormal, and the top 10\% holds 65.6\%, 74.9\%, and 66.8\%.
\item \textbf{Overlap of top neurons across datasets.} The Jaccard coefficient between the
100 neurons with the highest loading on $\mathbf{u}_1$ of two datasets (lower triangle of
\cref{tab:app_cross}).
\item \textbf{Direction similarity across datasets.} The $\mathbf{u}_1$ of each dataset is
mapped back to the full neuron coordinates (unselected neurons set to zero, the diagonal
calibration undone as in \cref{app:invariance}), and pairwise cosines are computed (upper
triangle of \cref{tab:app_cross}).
\end{itemize}

\begin{table}[!ht]
\centering
\caption{Comparison of $\mathbf{u}_1$ across datasets (Qwen3.5-2B). Upper triangle: cosine
similarity. Lower triangle: Jaccard coefficient of the top-100 neurons.}
\label{tab:app_cross}
\setlength{\tabcolsep}{4pt}
\renewcommand{\arraystretch}{1.12}
\setlength{\aboverulesep}{0pt}
\setlength{\belowrulesep}{0pt}
\resizebox{\columnwidth}{!}{%
\begin{tabular}{l|ccc}
\toprule
\rowcolor{vadHeader}
& \textbf{UCF-Crime} & \textbf{UBnormal} & \textbf{XD-Violence} \\
\midrule
UCF-Crime   & --    & 0.061 & 0.091 \\
UBnormal    & 0.064 & --    & $-$0.011 \\
XD-Violence & 0.093 & 0.105 & -- \\
\bottomrule
\end{tabular}%
}
\end{table}

For reference, the cosine of two random directions has a standard deviation of about
$1/\sqrt{D}\approx0.013$ ($D=6144$). Two random sets of 100 neurons out of $D$ overlap in
about 1.6 neurons in expectation, a Jaccard coefficient of about 0.008. The top-100 Jaccard
coefficients of 0.064--0.105 are 8--13 times this level. The cosines between UCF-Crime and
the other two datasets (0.061 and 0.091) are about 5--7 standard deviations of the random
level. The $\mathbf{u}_1$ of different datasets thus share high-loading neurons above chance,
while each dataset's event direction is mostly its own with a small shared part.

These statistics describe joint directions of neuron populations. They do not mean that a
single neuron corresponds to a single concept, or that neuron coordinates are shared across
backbones.


\end{document}